\documentclass[journal]{IEEEtran}
\usepackage{amsmath,amsfonts}
\usepackage{algorithmic}
\usepackage{algorithm}
\usepackage{array}
\usepackage{tabularx}
\usepackage{booktabs}
\usepackage{adjustbox}
\usepackage{makecell}
\usepackage{multirow}
\usepackage{threeparttable}
\usepackage[table,xcdraw]{xcolor}  
\usepackage{arydshln}  
\usepackage[caption=false,font=normalsize,labelfont=sf,textfont=sf]{subfig}
\usepackage{textcomp}
\usepackage{stfloats}
\usepackage{url}
\usepackage{verbatim}
\usepackage{graphicx}
\usepackage{cite}
\usepackage{amssymb}
\usepackage[pagebackref=true,breaklinks=true, colorlinks, bookmarks=false, citecolor=blue]{hyperref} 
\begin{document}

\title{Spectral-Adaptive Modulation Networks\\ for Visual Perception}

\author{Guhnoo Yun,~\IEEEmembership{Student Member, IEEE},
        Juhan Yoo, 
        Kijung Kim, 
        Jeongho Lee, \\ 
        Paul Hongsuck Seo, 
        Dong Hwan Kim,~\IEEEmembership{Member, IEEE}

\thanks{Manuscript received Month Day, Year; revised Month Day, Year.}
\thanks{Guhnoo Yun, Kijung Kim, Jeongho Lee, and Dong Hwan Kim 
          are with Korea University (KU) in Seoul, Korea, 
          and Korea Institute of Science and Technology (KIST) in Seoul, Korea. 
          (e-mail: \{doranlyong, plan100day, kape67, gregorykim\}@kist.re.kr).}
\thanks{Juhan Yoo is with Dong-A University in Busan, Korea (e-mail: unchinto@dau.ac.kr).}
\thanks{Paul Hongsuck Seo is with Korea University in Seoul, Korea (e-mail: phseo@korea.ac.kr).}
\thanks{Corresponding author: Dong Hwan Kim (e-mail: gregorykim@kist.re.kr).}
\thanks{Accepted for publication in IEEE TPAMI}
\thanks{Our code is available at \url{https://github.com/DoranLyong/SPANetV2-official}.}
}

\markboth{Journal of \LaTeX\ Class Files,~Vol.~14, No.~8, August~2021}%
{Shell \MakeLowercase{\textit{et al.}}: A Sample Article Using IEEEtran.cls for IEEE Journals}

\IEEEpubid{0162--8828~\copyright~2026 IEEE. Personal use is permitted, but
republication/redistribution requires IEEE permission. DOI: 10.1109/TPAMI.2026.3690455}

\maketitle

\begin{abstract}
Recent studies have shown that 2D convolution and self-attention exhibit distinct spectral behaviors, and optimizing their spectral properties can enhance vision model performance. However, theoretical analyses remain limited in explaining why 2D convolution is more effective in high-pass filtering than self-attention and why larger kernels favor shape bias, akin to self-attention. In this paper, we employ graph spectral analysis to theoretically simulate and compare the frequency responses of 2D convolution and self-attention within a unified framework. Our results corroborate previous empirical findings and reveal that node connectivity, modulated by window size, is a key factor in shaping spectral functions. Leveraging this insight, we introduce a \textit{spectral-adaptive modulation} (SPAM) mixer, which processes visual features in a spectral-adaptive manner using multi-scale convolutional kernels and a spectral re-scaling mechanism to refine spectral components. Based on SPAM, we develop SPANetV2 as a novel vision backbone. Extensive experiments demonstrate that SPANetV2 outperforms state-of-the-art models across multiple vision tasks, including ImageNet-1K classification, COCO object detection, and ADE20K semantic segmentation.

\end{abstract}

\begin{IEEEkeywords}
SPANetV2, deep learning, visual perception, convolution, self-attention, spectral-adaptive modulation, neural networks.
\end{IEEEkeywords}

\section{Introduction}
\label{sec:introdution}
Developing powerful backbone architectures is important in computer vision, and Convolutional Neural Networks (CNNs) have long been the predominant approach for addressing a wide range of vision tasks, due to their strong inductive bias in capturing local information~\cite{he2016deep, krizhevsky2017imagenet, simonyan2014very, szegedy2015going}. Over the last few years, however, with the success of Vision Transformer (ViT)~\cite{dosovitskiy2020image}, many follow-up models based on the ViT paradigm have gained significant traction, and have become one of the dominant structures as an alternative to CNNs in various computer vision tasks including image classification~\cite{touvron2021training, wu2021cvt, liu2021swin, vaswani2021scaling}, object detection~\cite{carion2020end, zhu2020deformable, zheng2020end}, segmentation~\cite{wang2021max, wang2021end}, and beyond~\cite{chang2021augmented, neimark2021video, wang2021transformer, gong2022future}. The primary reason for ViT's success lies in the self-attention mechanism, which effectively captures long-range dependencies without relying on a strong inductive bias~\cite{dosovitskiy2020image, naseer2021intriguing, tuli2021convolutional, yu2021rethinking, mao2022towards, chu2021Twins}. This widely accepted notion has prompted the development of numerous variants of self-attention, aimed at further improving the performance of ViT-based models~\cite{touvron2021training, yuan2021tokens, wang2021pyramid, liu2021swin, dong2022cswin}. 

\IEEEpubidadjcol

Following the success of ViTs, many researchers expected Transformers to dominate computer vision. However, recent studies challenge this view, demonstrating competitive performance without self-attention mechanisms. Notably, Multi-Layer Perceptron (MLP)-based approaches ~\cite{tolstikhin2021mixer, hou2022vision, touvron2022resmlp, tang2022image, chen2023cyclemlp} have shown promising results in vision tasks, narrowing the performance gap between MLP-based models and ViTs. Beyond MLPs, various methods have also emerged as effective alternatives to self-attention. For instance, Fast Fourier Transform (FFT)-based models ~\cite{lee-thorp-etal-2022-fnet, rao2023gfnet, tatsunami2024fft} have demonstrated feasibility as self-attention substitutes. Similarly, graph convolution ~\cite{han2022vision} and clustering-based token-mixers~\cite{ma2023image, liang2023clusterfomer} have been effectively employed to encode image patches. These approaches highlight that self-attention is not the sole dominant method for spatial patch modulation. Instead, diverse techniques can enhance vision backbones.

Concurrently, CNN research has evolved by integrating strategies inspired by ViTs, revitalizing its significance. ResNet Strikes Back ~\cite{wightman2021resnet} demonstrated that applying advanced training techniques from ViTs, such as those in DeiT ~\cite{touvron2021training} and Swin Transformer~\cite{liu2021swin}, improves ResNet-50 performance by $2.3\%$. Similarly, ConvNeXt~\cite{liu2022convnet} incorporated design elements from Swin models~\cite{liu2021swin}, showing that CNNs can rival ViTs by adopting larger kernels and advanced block modules. Numerous other studies further support these findings~\cite{ding2022scaling, liu2023more, hou2024conv2former, rao2022hornet, wang2023internimage, yu2023metaformer}.

Several studies have explored the fundamental differences between self-attention and convolution ~\cite{tuli2021convolutional, dosovitskiy2020image, yu2021rethinking, mao2022towards}, leveraging their distinct advantages ~\cite{li2023uniformer, dai2021coatnet, zhao2021battle, zhang2023vitaev2, pan2022fast}. Recent research has moved beyond inductive bias analysis to evaluating effectiveness in the spectral domain. A common finding ~\cite{park2022how, bai2022improving, wang2022antioversmooth, wang2020high, wang2022vtc} is that self-attention acts as a low-pass filter, whereas convolution effectively captures high-frequency components. Notably, Wang~\textit{et al.} ~\cite{wang2022antioversmooth} proposed a framework that explains why self-attention behaves as a low-pass filter in the Fourier domain. In contrast, the spectral properties of CNNs have primarily been examined through empirical observations~\cite{park2022how, bai2022improving, wang2020high, chen2024revealing}. Thus, a theoretical framework is needed to characterize the origins of these spectral differences.

To address this, we draw inspiration from kernel size effects. Ding~\textit{et al.}~\cite{ding2022scaling} argue that large kernel design expands effective receptive fields (ERFs) and favors shape bias, akin to self-attention. Intuitively, enlarging 2D convolution kernels broadens spatial interactions between patches, analogous to expanding node connectivity in graphs. In contrast, self-attention can be represented as a fully connected graph, where patches act as nodes and attention weights define edges representing their interactions. Given these perspectives, we adopt a graph-based framework to represent both 2D convolution and self-attention in Euclidean space and analyze them using graph spectral analysis. As a result, we theoretically find that \textit{large-kernel convolution functions as a low-pass filter, akin to self-attention, whereas small-kernel convolution inherently filters both low- and mid/high-frequency components.}

Building on these findings, we propose a novel token mixer, \textit{Spectral-Adaptive Modulation} (SPAM), for image feature aggregation. In our previous work~\cite{yun2023spanet}, it was demonstrated that \textit{optimally aggregating spectral components of image patches enhances the performance of vision models}. Building on this idea, our new SPAM encodes image patches in a spectral-adaptive manner by employing convolutions with multiple kernel sizes and applying frequency-domain mask filtering to modulate spectral characteristics. Furthermore, we introduce \textit{SPANetV2}, incorporating SPAM within the MetaFormer architecture~\cite{yu2023metaformer}.  SPANetV2 is evaluated on multiple vision tasks, including image classification, object detection, and semantic segmentation, achieving superior performance over previous state-of-the-art models.

Our contributions are threefold. (1) We theoretically analyze convolution and self-attention through graph signal processing, providing a unified framework for spectral analysis from a graph perspective. This view enables a comprehensive comparison of their spectral properties, explaining why small-kernel designs are more adept at high-pass filtering, whereas large-kernel designs favor low-pass filtering, akin to self-attention, corroborating previous experimental findings~\cite{park2022how, bai2022improving, wang2020high, ding2022scaling}. (2) Building on our previous work, we propose SPAM, a novel token-mixer for spectral-adaptive image feature aggregation, leveraging multi-kernel convolutions and frequency-domain filtering. We then introduce SPANetV2, a new vision architecture that incorporates SPAM. (3) Our SPANetV2 is evaluated on multiple vision tasks, including image classification~\cite{deng2009imagenet}, object detection~\cite{lin2014microsoft}, and semantic segmentation~\cite{zhou2017scene}. Our results demonstrate that SPANetV2 outperforms state-of-the-art models.


\section{Related Works}
\label{sec:relatedworks}
\subsection{ CNNs, ViTs, and Beyond for Vision Architecture}
CNNs~\cite{he2016deep, krizhevsky2017imagenet, simonyan2014very, szegedy2015going} predominated vision tasks until the emergence of Transformers~\cite{vaswani2017attention}, initially designed to capture long-range dependencies in language tasks via self-attention. Inspired by their success, Transformers were introduced into vision by ViT~\cite{dosovitskiy2020image}, achieving remarkable performance in image classification. Subsequent studies have improved ViT's self-attention mechanisms using various approaches, including shifted windows~\cite{liu2021swin}, relative positional encoding~\cite{wu2021rethinking}, anti-aliasing attention~\cite{qian2021blending}, and convolution integration~\cite{d2021convit, guo2022cmt, wu2021cvt}.

While self-attention is widely considered crucial for ViT performance, several studies have questioned its necessity. For instance, fully replacing self-attention with MLP-based mixers can yield competitive results~\cite{tolstikhin2021mixer, touvron2022resmlp}. These findings have prompted further exploration of alternative token mixers~\cite{chen2022cyclemlp, hou2022vision}, challenging attention dominance through various non-attention approaches~\cite{martins2020sparse, martins2022infinite, lee-thorp-etal-2022-fnet, rao2023gfnet, tatsunami2024fft, han2022vision, ma2023image, liang2023clusterfomer}.

Several other studies have explored transformer architectures within meta-architectures by replacing self-attention with non-parametric token mixers. For example, ShiftViT~\cite{wang2022shift} employs a partial shift operation~\cite{lin2019tsm}, and PoolFormer~\cite{yu2022metaformer,yu2023metaformer} utilizes spatial average pooling. Both achieve competitive performance across various vision tasks, highlighting the effectiveness of meta-architecture block and stage designs. Following these, we adopt the MetaFormer architecture ~\cite{yu2023metaformer}, replacing the token mixer with our proposed SPAM.

\subsection{Frequency in Images}
In computer vision literature, images have been extensively studied in the spectral domain. Normally, low frequencies encode global structures and color information, while high frequencies capture fine details (\textit{e.g.}, local edges/textures)~\cite{chen2019drop, cooley1969fast, deng1993adaptive}. Several studies~\cite{park2022how, bai2022improving, wang2022antioversmooth, wang2020high, wang2022vtc} indicate that self-attention predominantly captures low-frequency representations but struggles with high-frequency information. In contrast, convolution exhibits the opposite trend. Consistent with these findings, several works~\cite{pan2022fast, bai2022improving, huang2023adaptive,rao2023gfnet, tatsunami2024fft, yun2023spanet} show that optimizing frequency components for feature aggregation enhances vision model performance. 

Wang~\textit{et al.} ~\cite{wang2022antioversmooth} proposed a theoretical framework for analyzing ViT features in the Fourier domain. However, to the best of our knowledge, the spectral properties of convolutions in CNNs have primarily been examined through empirical methods, such as spectral noise testing ~\cite{bai2022improving, wang2020high, chen2024revealing} and observation of spectral components ~\cite{park2022how} in pre-trained models. In this paper, we address this gap by introducing a unified framework using graph spectral analysis. Since images can be represented as grid-structured graphs, they can be analyzed from a graph perspective.

\subsection{Frequency in Graphs}
Graph signal processing (GSP) theory~\cite{shuman2013emerging} extends signal processing to graph-based data, employing the graph Fourier transform (GFT) to analyze and manipulate signals on irregular structures. By leveraging graph topology, GSP provides powerful tools for applications such as network analysis and image processing. Balcilar~\textit{et al.}~\cite{balcilar2020analyzing} proposed a theoretical framework for understanding graph neural networks (GNNs) through frequency response analysis. 

Building on this framework, we analyze how convolutional kernels of varying sizes capture spectral components, offering theoretical insights into the spectral differences between convolution and self-attention. Additionally, we explain why larger kernels favor shape bias, akin to self-attention in vision tasks~\cite{ding2022scaling}. Expanding on this analysis, we propose the SPAM mixer for spectral-adaptive visual feature aggregation and introduce SPANetV2 as a new vision backbone.

\section{Spectral Analysis in A Graph Formulation}
\label{sec:spectral_analysis}
This section compares 2D convolution in CNNs with self-attention in ViTs through spectral analysis via graph spectral filtering~\cite{chung1997spectral, shuman2013emerging, balcilar2020analyzing}. Balcilar~\textit{et al.}~\cite{balcilar2020analyzing} proposed a framework that generalizes non-Euclidean graph convolution, 2D Euclidean convolution, and graph attention. Modeling images as a regular grid graph allows convolution and self-attention to be reformulated within this framework, enabling their spectral comparison.

\subsection{Background}

\subsubsection{Spectral filtering in graphs}
Let $G$ represent an undirected graph consisting of $N$ nodes and a variable number of edges. The structural properties of the graph are given by the adjacency matrix $A \in \{0,1\} ^{N \times N}$, and the normalized graph Laplacian $L$ can be defined as $L=I-D^{-1/2}AD^{-1/2}$. Here, $D \in \mathbb{R}^{N \times N}$ is the diagonal degree matrix and $I$ is the identity matrix. $L$ can be expressed using eigendecomposition as $L = U \textrm{diag}(\boldsymbol{\lambda})U^{T}$, where $U \in \mathbb{R}^{N \times N}$ denotes a matrix whose columns are the eigenvectors. $\textrm{diag}(\boldsymbol{\lambda})$ generates a diagonal matrix whose diagonal components are derived from the elements of eigenvalues $\boldsymbol{\lambda}\in \mathbb{R}^{N}$. Since $L$ is a real symmetric matrix, it has a complete set of orthonormal eigenvectors, and its eigenvalues are constrained to the range between 0 and 2~\cite{chung1997spectral}. In graph signal processing theory~\cite{shuman2013emerging}, graph signals can be filtered in the spectral domain by utilizing the eigenvectors of the normalized graph Laplacian matrix as the bases in GFT. Given a signal $\textbf{h} \in \mathbb{R}^{N}$, the GFT is defined as $\hat{\textbf{h}}=U^{T}\textbf{h}$, and the inverse GFT (iGFT) is $\textbf{h}=U\hat{\textbf{h}}$. Then, the convolution $\ast$ between the signal $\textbf{h}$ and convolution kernel $\textbf{f}$ is described as follows:
\begin{equation} \label{eq_gsf}
    \textbf{f} \ast \textbf{h} = U((U^{T} \textbf{f}) \odot (U^{T} \textbf{h})) = U\textrm{diag}(\Phi (\boldsymbol{\lambda}))U^{T}\textbf{h},
\end{equation}
where $\odot$ denotes the element-wise product, $\Phi(\cdot)$ is the desired filter function, and $\textrm{diag}(\Phi (\boldsymbol{\lambda}))$ represents the convolution kernel in the spectral domain.

\subsubsection{Frequency response of GNNs}
GNN operation involves two steps: aggregating neighborhood nodes and updating through linear transformation. For spatial GNNs, they can be generalized as propagating node features to neighboring nodes~\cite{balcilar2020analyzing}, as shown below:
\begin{equation} \label{eq_spatial_gnn_propa}
H^{(l+1)} = \sigma(\sum_{i=1}^{N_{f}} C^{(i)} H^{(l)} W^{(l,i)}),
\end{equation}
where $\sigma$ is an activation function, $N_{f}$ is the number of filters, $C^{(i)}\in \mathbb{R}^{N \times N}$ is a convolution support for the $i$-th filter that defines how the node features are propagated to neighborhood nodes, $H\in \mathbb{R}^{N \times d_{l}}$ is a set of $N$ node features with $d$ dimensions, and $W^{(l,i)}\in \mathbb{R}^{d_{l}\times d_{l+1}}$ is a trainable matrix for the $i$-th filter of the $l$-th layer. Spectral GNNs work all the same way as spatial GNNs, but are defined using a function of eigenvalues $\Phi_{i} (\boldsymbol{\lambda})$ and eigenvectors $U$ of the normalized graph Laplacian $L$. Specifically, $C^{(i)}$ can be interpreted in the spectral domain and is derived by $U\textrm{diag}(\Phi_{i} (\boldsymbol{\lambda}))U^{T}$~\cite{balcilar2020analyzing}. So, the frequency response of $C^{(i)}$ is defined as:
\begin{equation} \label{eq_freq_response}
\Phi_{i} (\boldsymbol{\lambda}) = \textrm{diag}^{-1}(U^{T} C^{(i)} U),
\end{equation}
where $\textrm{diag}^{-1}(\cdot)$ returns the vector consisting of the diagonal elements of the given matrix.

\subsection{Graph Formulation for Convolution and Self-Attention}
The 1D sequential inputs of ViTs and the 2D regular grid inputs of CNNs are special cases of graphs. Thus, both convolution and self-attention can be analyzed within the GNN framework. For clarity, we define image features as $\widetilde{X}\in \mathbb{R}^{D \times H \times W}$, where $D$, $H$, and $W$ denote the number of feature dimensions, height, and width, respectively. The $d$-th feature is represented as $\widetilde{X}_{d}\in \mathbb{R}^{H \times W}$.

\begin{figure*}[t]
\centering
\subfloat[]{\includegraphics[width=0.46\textwidth]{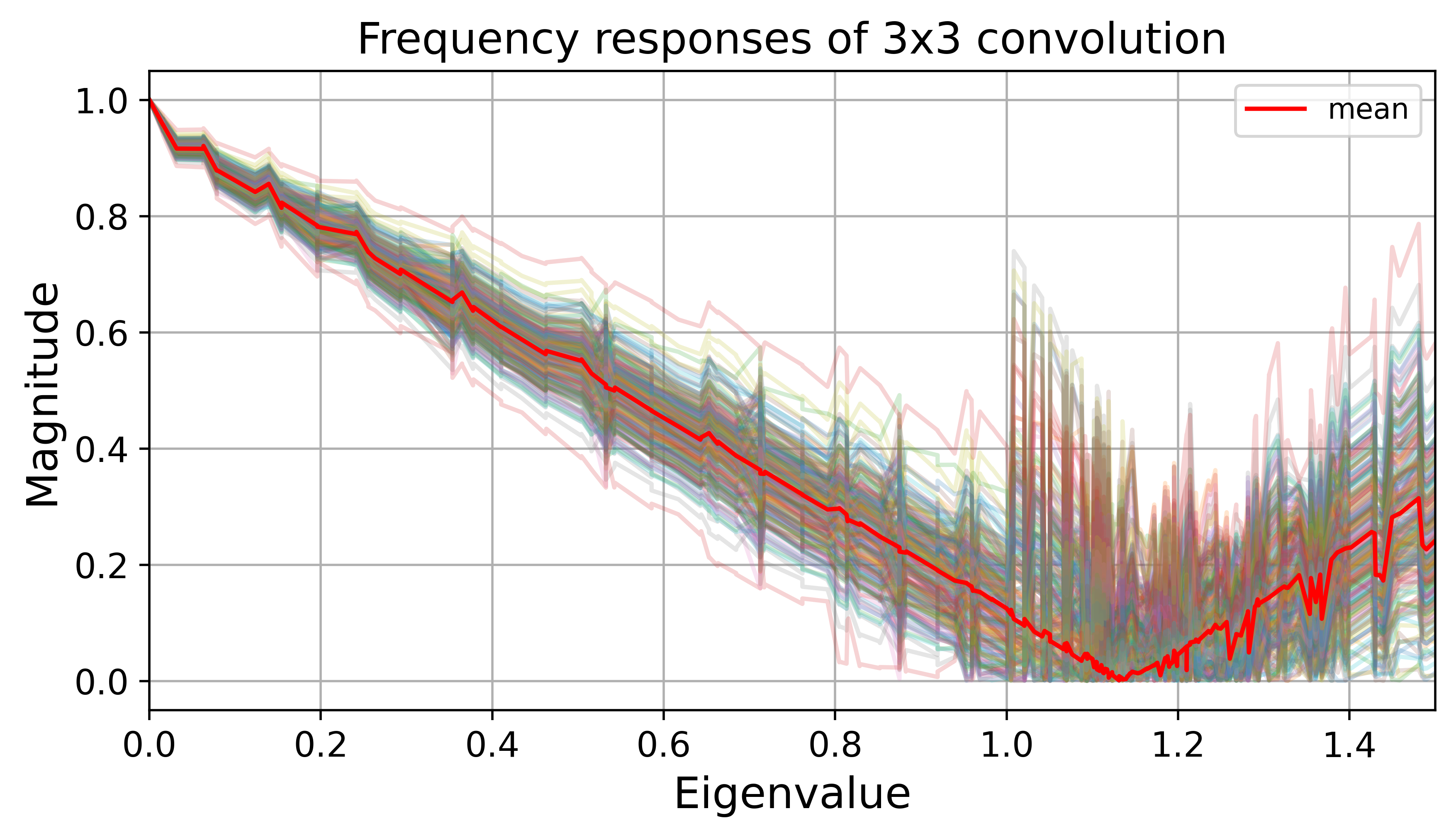}} \quad
\subfloat[]{\includegraphics[width=0.46\textwidth]{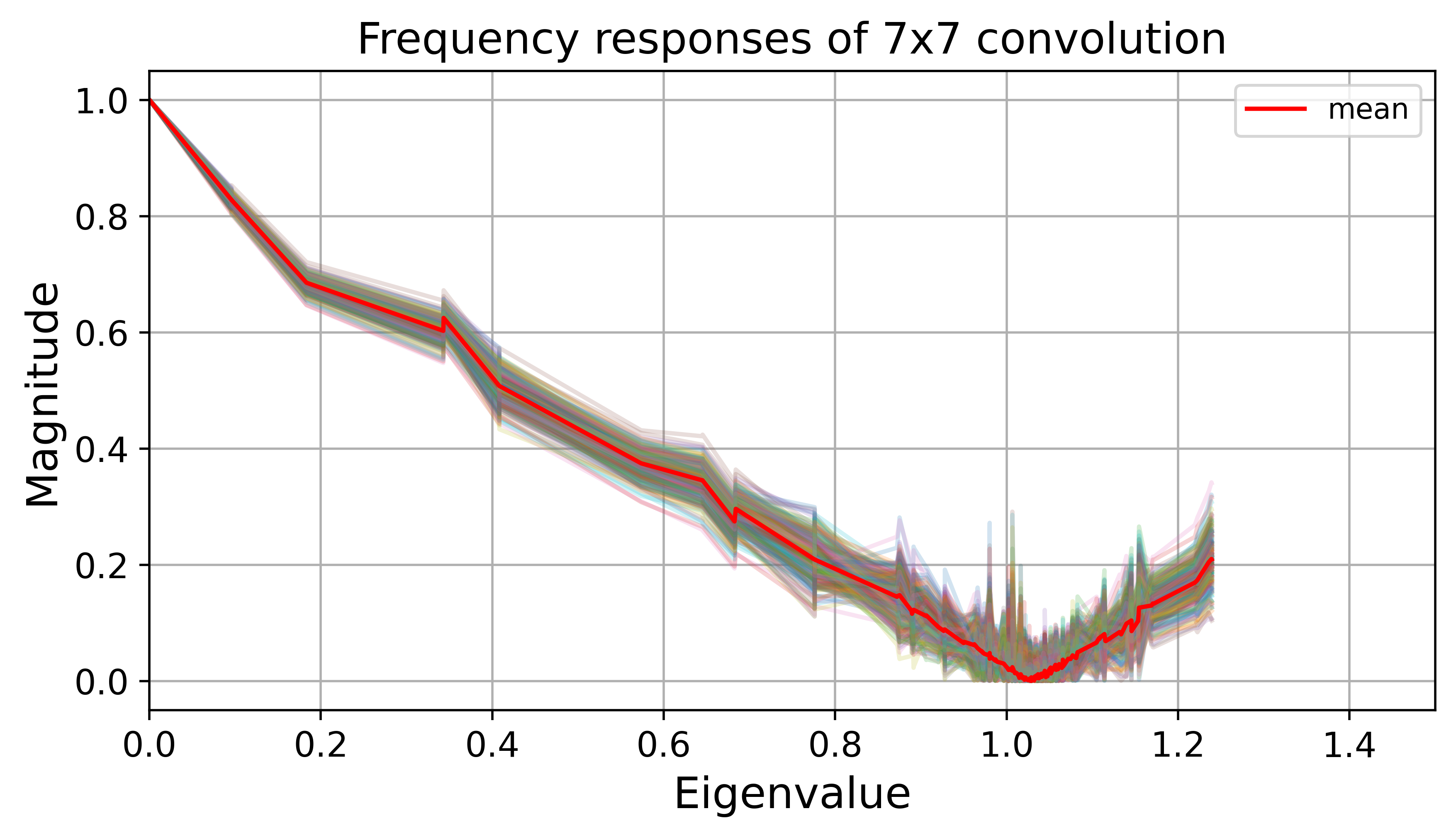}} \\ [-1em]  

\subfloat[]{\includegraphics[width=0.46\textwidth]{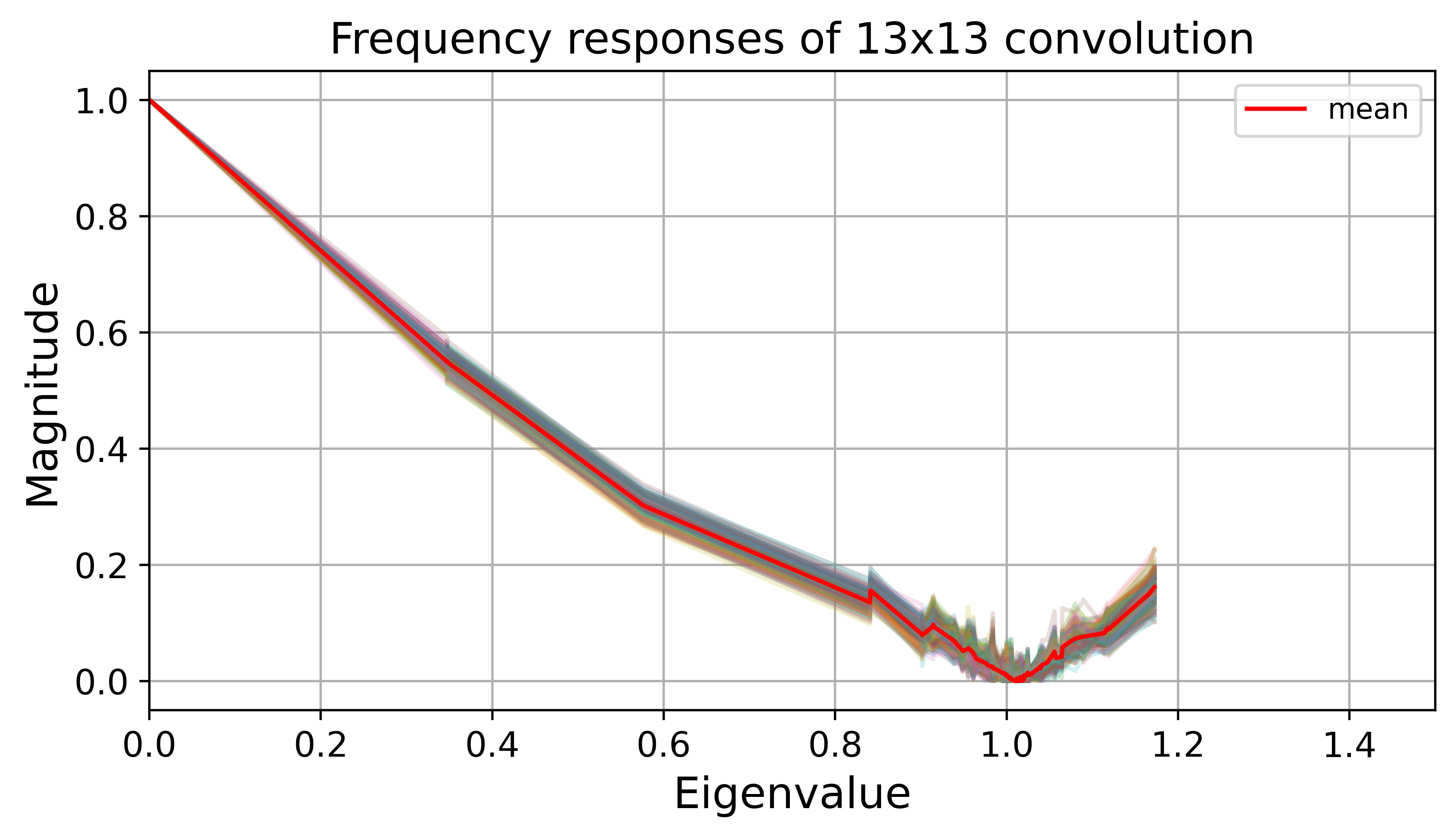}} \quad
\subfloat[]{\includegraphics[width=0.46\textwidth]{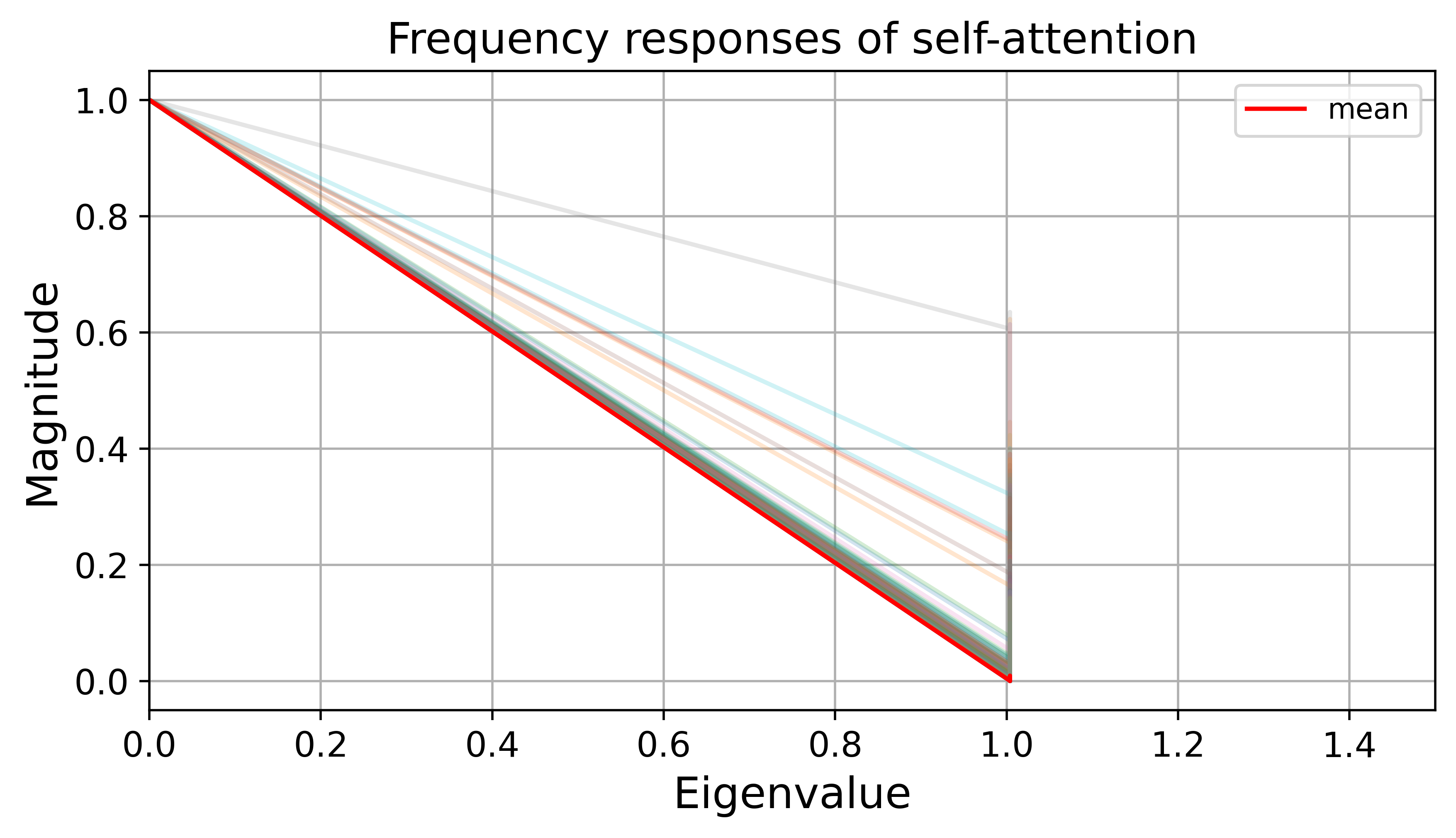}}
\caption{\textbf{Simulation examples of frequency response}. (a)-(c) show responses of 2D Euclidean convolutions with increasing kernel sizes, and (d) shows responses of self-attention. All responses are obtained with random weights. The input patch size is set to $16 \times 16$, inspired by ViT~\cite{dosovitskiy2020image}. As the convolution kernel size increases, the cut-off frequency shifts closer to one, making it behave more like a low-pass filter, akin to self-attention.}
\label{fig_freq_response}
\end{figure*}

\subsubsection{2D convolution using a sparse matrix}
A 2D Euclidean convolution with a kernel $\widetilde{K} \in \mathbb{R}^{m\times m}$ is performed by sliding the kernel over an image feature $\widetilde{X}_{d}$, applying element-wise multiplication to each grid patch, and summing the results to generate the output feature. This section formulates this operation as sparse matrix multiplication~\cite{liu2015sparse, tan2022accelerating}, with a focus on constructing the sparse convolution support matrix $C^{(i)}$. For simplicity, we consider a single-channel 2D convolution and hence denote $C^{(i)}$ simply as $C$.

To formulate 2D convolution as a sparse matrix multiplication, we define a sparse matrix $C \in \mathbb{R}^{H_{out} W_{out} \times HW}$. Given the $z$-th kernel element located at position $(r,t)$ in the kernel $\widetilde{K}$, indexed using zero-based ordering as $z=r\times m+t$, the matrix $C$ is constructed as a weighted sum of fixed basis convolution support matrices $B^{(z)}\in \{0, 1\}^{H_{out} W_{out} \times HW}$:
\begin{equation}
    C = \sum_{z=0}^{Z-1} k^{(z)} B^{(z)},
\end{equation}
where $Z = m \times m$ and $k^{(z)}\in \mathbb{R}$ denotes the kernel value at index $z$. Each $B^{(z)}$ encodes the spatial relationship between the $z$-th kernel element position and the corresponding input positions across all sliding windows, accounting for zero-padding, by indicating which input elements are multiplied by the $z$-th kernel weight at each location. The output dimensions are defined as: 
\begin{subequations}
\begin{align}
H_{out} &= \left \lfloor \frac{H + 2p - m}{s}\right \rfloor + 1, \\
W_{out} &= \left \lfloor \frac{W + 2p - m}{s}\right \rfloor + 1,
\end{align}
\end{subequations}
where $\lfloor \cdot \rfloor$ is the floor function, $p$ is the padding size, $m$ is the kernel size, and $s$ is stride. We set $p=\left \lfloor (m-1)/2 \right \rfloor$ and $s=1$ to maintain the output size equal to the input size.

To construct the elements of $B^{(z)}$, we consider the position $(r, t)$ of the $z$-th kernel element. For an output position $(y, x)$, the corresponding indices in the flattened output and input are computed as:
\begin{subequations}
\begin{align}
a &= y \times W + x, \\
b &= (y + r - p) \times W + (x + t - p),
\end{align}
\end{subequations}
where $a$ and $b$ denote the indices in the flattened output and input, respectively. The input index $b$ is adjusted for padding and aligned with the corresponding kernel element. The elements of $B^{(z)}$ are then assigned as:
\begin{equation} \label{eq_conv_support}
B^{(z)}_{a,b}= B^{(r,t)}_{a,b} = \
\begin{cases} 
1 & \text{if } 0 \leq (y + r - p) < H \text{ and } \\ 
  & \quad  0 \leq (x + t - p) < W, \\
0 & \text{otherwise}.
\end{cases}
\end{equation}
This condition ensures that $B^{(z)}$ correctly maps between the kernel element $k^{(z)}$ and the valid regions of the input feature. Specifically, $B^{(z)}_{a,b} = 1$ if the kernel element $k^{(z)}$ lies within the receptive field of $\widetilde{X}_{d}$, including zero-padding; otherwise, $B^{(z)}_{a,b} = 0$. Thus, the 2D convolution operation is formulated as:
\begin{equation} \label{eq_gen_conv2d}
\widetilde{X}_{d} \circledast \widetilde{K} = CX_{d} = \sum_{z=0}^{Z-1} B^{(z)} X_{d} k^{(z)},
\end{equation}
where $X_{d} \in \mathbb{R}^{HW}$ represents the flattened $\widetilde{X}_{d}$. For clarity and concise formulation, the reshaping step to align tensors on the left-hand and right-hand sides is omitted. This formulation represents 2D convolution as sparse matrix multiplication, emphasizing its relationship with the spatial structure of the input. For simplicity, the bias term is omitted. This sparse matrix representation enables frequency response analysis of 2D convolution by examining the properties of $C$, which directly stem from the convolution kernel $\widetilde{K}$.

\subsubsection{Self-attention}
Self-attention in ViTs models relationships between patches, representing each patch as a node in a fully connected graph. The connection strengths are determined by an attention weight matrix, computed from the query $(Q)$ and key $(K)$ matrices. Given $Q, K \in \mathbb{R}^{HW \times d_{h}}$, obtained by projecting $X \in \mathbb{R}^{D \times HW}$ with trainable weight matrices, the attention matrix $E \in \mathbb{R}^{HW \times HW}$ is defined as:
\begin{equation} \label{eq_att_score} E = \text{softmax}\left(\frac{Q K^{T}}{\sqrt{d_{h}}}\right), \end{equation}
where $d_{h}$ denotes the head dimension, and $\text{softmax}$ is applied along the last dimension to normalize attention scores.

The final attention mechanism is expressed as $E(X^{T}W_{v})=EX^{T}W_{v}$, where $W_{v} \in \mathbb{R}^{D \times d_{h}}$ represents the value projection matrix. This can also be reformulated using convolution support matrices by defining $C^{(i)} \in \mathbb{R}^{HW \times HW}$ as the attention weights for each input patch pair. Unlike the binary values in ~\eqref{eq_conv_support}, $C^{(i)}$ contains continuous attention weights between query-key pairs:
\begin{equation} \label{eq_att_score_conv_support} C^{(i)}_{a,b} = \frac{\text{exp} ( Q_{a} K^{T}_{b} / \sqrt{d_{h}} )}{\sum_{j=0}^{J-1} \text{exp}(Q_{a} K^{T}_{j} / \sqrt{d_{h}})}, \end{equation}
where $C^{(i)}_{a,b} \in \mathbb{R}$ denotes the attention weight between the $a$-th query and $b$-th key, and $J = H \times W$. Consequently, self-attention can be rewritten as:
\begin{equation} \label{eq_att_conv_support} \textit{Attention}(X) = E(X^{T}W_{v}) = \sum_{i=1}^{d_{h}} C^{(i)} X^{T} w^{(i)}_{v}, \end{equation}
where $w^{(i)}_{v} \in \mathbb{R}^{D \times d_{h}}$ is a sparse matrix with only its $i$-th column containing the corresponding column vector from $W_{v}$.

\subsection{Frequency Response Profiling}

\subsubsection{Simulation}
For straightforward profiling~\cite{balcilar2020analyzing}, we run 240 simulations with both randomly generated weights and $16 \times 16$ patches to estimate the expected frequency responses of $C^{(i)}$ for 2D convolution and self-attention. In 2D convolution, patch connections are dictated by the kernel size, forming a locally connected graph, whereas self-attention constructs a fully connected graph, linking all patches. Based on these properties, we construct adjacency matrices, compute their eigenvectors, and simulate frequency responses using ~\eqref{eq_freq_response}. Fig.~\ref{fig_freq_response} shows the simulated frequency responses.

As can be seen, small kernels tend to filter both low- and slightly mid/high-frequency components, while larger kernels shift the cut-off frequency closer to one, functioning as low-pass filters similar to self-attention. This trend occurs because a larger receptive field averages out high-frequency components. Although these results are simulated rather than learned, they demonstrate distinct different spectral behaviors in convolution and self-attention, as observed in prior studies ~\cite{park2022how, bai2022improving, wang2022antioversmooth, wang2020high, wang2022vtc}.

\subsubsection{Discussion}
The graph spectral profile range is determined by the node connectivity and computed via the eigendecomposition of the normalized graph Laplacian $L$. In 2D Euclidean convolution, image patches are represented as nodes in a grid graph, where connections are dictated by the kernel size. For example, regular $3\times 3$ kernels link each node to its eight neighbors, resulting in a sparse adjacency matrix $A$ that captures only local connections. Thus, $L$ for the 2D convolution encodes the local structure of these connections. Following ~\eqref{eq_gsf}, ~\eqref{eq_spatial_gnn_propa}, and ~\eqref{eq_gen_conv2d}, the 2D Euclidean convolution operation is expressed in the graph spectral domain as: 
\begin{equation} \label{eq_conv2d_sepctral}
\widetilde{X}_{d} \circledast \widetilde{K} = U f(\Lambda)U^{T}X_{d},
\end{equation}
where $f(\Lambda)$ modifies the eigenvalues based on the kernel $\widetilde{K}$, and $\Lambda$ is a diagonal matrix whose diagonal components are eigenvalues of $L$.

Based on the simulation results in Fig.~\ref{fig_freq_response}, node connectivity can be considered a key factor in utilizing the spectral characteristics of convolution and self-attention. For instance, Swin Transformer~\cite{liu2021swin} captures high-frequency signals more effectively than ViT~\cite{dosovitskiy2020image} by employing the shifted window mechanism~\cite{yun2023spanet}, which enhances local node connectivity in self-attention. Similarly, HiLo attention~\cite{pan2022hilo} uses a local window to improve high-frequency capture. Conversely, larger kernels increase global node connectivity, acting as low-pass filters. Low-frequency components encode global structures associated with shape bias~\cite{subramanian2023spatial}, aligning with Ding~\textit{et al.}~\cite{ding2022scaling}, who found that large kernel designs favor shape bias similar to self-attention. These insights suggest that adjusting kernel size can modulate spectral properties in convolutional operations. Furthermore, kernel size modulation can induce spectral functions akin to self-attention, and \textit{vice versa}. In the next section, we introduce a novel convolutional mixer that aggregates visual features in a spectral-adaptive manner by leveraging this property.

\section{A Spectral-Adaptive Token Mixer}
\label{sec:SPAM}
From a spectral perspective, optimal aggregation of spectral components in image patches enhances vision model performance~\cite{yun2023spanet}. In this section, we propose a novel token mixer incorporating spectral-adaptive convolutional modulation. 

\subsection{Convolutional Modulation}
Convolutional modulation~\cite{yang2022focal,yun2023spanet,hou2024conv2former} leverages convolution to mimic self-attention through a different way. This approach first aggregates contextual information, then modulates projected feature values via element-wise multiplication:
\begin{equation} \label{eq_conv_modulation} 
\widetilde{Y} = \mathcal{V}(\widetilde{X}) \odot \mathcal{C}(\widetilde{X}), \end{equation}
where $\mathcal{V}(\widetilde{X})$ represents projected feature values, $\mathcal{C}(\widetilde{X})$ denotes aggregated contexts, and $\widetilde{Y}\in \mathbb{R}^{D\times H\times W}$ is the refined representation. Here, $\mathcal{V}(\cdot)$ and $\mathcal{C}(\cdot)$ correspond to value projection and context aggregation functions, respectively. We define $\mathcal{V}(\cdot)$ as a non-linear transformation:
\begin{equation} \label{eq_proj_value} 
\mathcal{V}(\widetilde{X}) = \sigma(\text{Linear}(\widetilde{X}))\in \mathbb{R}^{D\times H \times W}, 
\end{equation}
where $\text{Linear}(\cdot)$ denotes a linear projection, and $\sigma(\cdot)$ represents an activation function. Next, we introduce $\mathcal{C}(\cdot)$ for spectral-adaptive context aggregation.

\begin{figure}[t] 
    \centering
    \includegraphics[width=\linewidth]{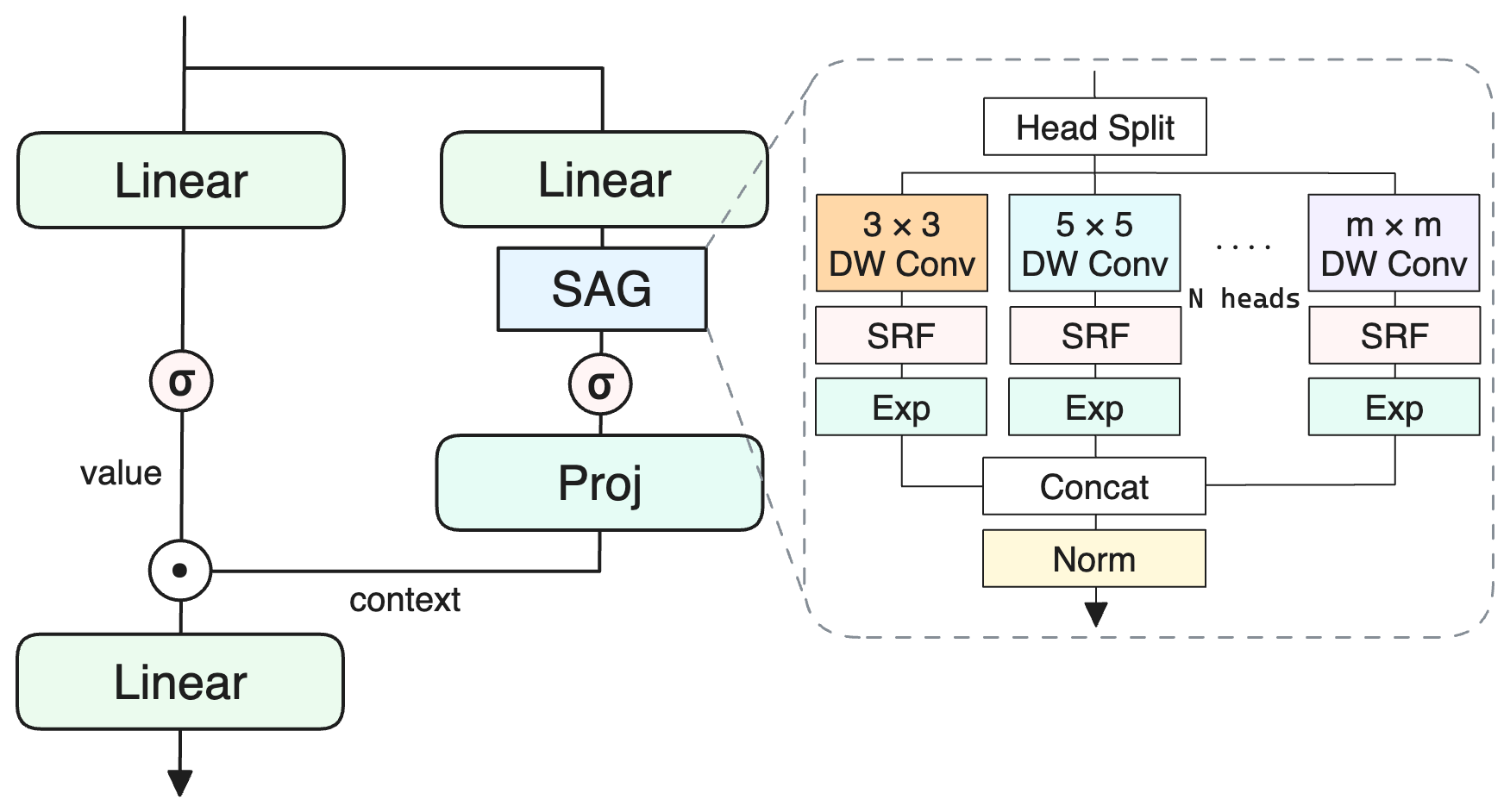} 
    \caption{\textbf{Overview of the SPAM mixer}. The \textit{Head Split} layer evenly partitions the input along feature dimensions based on the number of heads. \textit{DWConv} denotes depthwise convolution, while \textit{SRF} re-scales the spectral components of \textit{DWConv}'s output. All linear layers preserve input dimensions, except \text{Exp}, which doubles the feature dimensions, and \text{Proj}, which halves them.} 
    \label{fig_SPAM}
\end{figure}

\subsection{Spectral-Adaptive Context Aggregation}

To achieve spectral-adaptive context aggregation, we introduce the \textit{spectral-adaptive gate} (SAG), as illustrated in Fig.~\ref{fig_SPAM}. SAG comprise $N$ heads to capture diverse spectral responses, with each head processing an evenly partitioned input segment of size $d_{h}=\frac{D}{N}$ along the feature dimensions. The input is obtained through a linear transformation:
\begin{equation} \label{eq_proj_ctx} 
\widetilde{X}^{'} = \text{Linear}(\widetilde{X}) \in \mathbb{R}^{D\times H \times W}. 
\end{equation}
Referring to ~\eqref{eq_conv2d_sepctral}, we can decompose the spectral components of the input filtered by the depthwise convolution:
\begin{equation} \label{eq_spectrum}
\text{Conv}_{m\times m}(\widetilde{X}_{d}^{'}) =  Uf(\Lambda)\hat{X}_{d}^{'} 
 = \sum_{n=1}^{HW}f(\lambda_{n})\hat{x}^{'}_{dn}\mathbf{u}_{n}               
\end{equation} 
where $\text{Conv}_{m\times m}(\cdot)$ is the $m\times m$ depthwise convolution, $\hat{X}_{d}^{'}=U^{T}X_{d}^{'} \in \mathbb{R}^{HW}$ denotes the GFT, $\hat{x}^{'}_{dn}\in \mathbb{R}$ is its element, $f(\Lambda)$ is a diagonal matrix defining the filter function, $f(\lambda_{n})\in \mathbb{R}$ is its element, and $\mathbf{u}_{n}$ is a column vector of $U$. Since $\mathbf{u}_{n}$ is the Fourier basis in GFT, we can say that convolution scales and aggregates spectral components. However, as shown in Fig.~\ref{fig_freq_response}, fixed-size convolution constrains spectral modulation based on kernel size. To address this, we assign different kernel sizes to each head. 


The use of multiple kernels is a widely adopted design~\cite{yang2022focal, yu2024inceptionnext}. To enhance adaptability, we introduce the \textit{spectral re-scaling filter} (SRF),  which refines the filtered spectrum using learnable parameters. This modification is applied to ~\eqref{eq_spectrum} as follows:
\begin{equation} \label{eq_srf_1}
\text{SRF}(\text{Conv}_{m\times m}(\widetilde{X}_{d}^{'}))=\sum_{n=1}^{HW} \psi_{n} f(\lambda_{n})\hat{x}^{'}_{dn}\mathbf{u}_{n},              
\end{equation} 
where $\psi_{n}$ is a learnable parameter constrained to the range $[0,1]$. In 2D grid graphs (images) with periodic boundaries, the graph Laplacian eigenvectors align with the classical 2D discrete Fourier basis, differing only in basis ordering~\cite{ekambaram2014graph, ekambaram2013circulant}. Since our focus is solely on scaling each spectral component, we reformulate ~\eqref{eq_srf_1} as a mask filtering problem in the 2D FFT to enable efficient implementation using \texttt{torch.fft.fft2}:
\begin{subequations}\label{eq_srf_2}
\begin{align} 
\text{SRF}(\text{Conv}_{m\times m}(\widetilde{X}_{d}^{'})) &= \mathcal{D}^{-1}(\Psi \odot \mathcal{D}(\text{Conv}_{m\times m}(\widetilde{X}_{d}^{'}))), \\
\Psi &= \phi(M),              
\end{align} 
\end{subequations}
where $\mathcal{D}$ and $\mathcal{D}^{-1}$ denote the FFT and its inverse, respectively. $\Psi$ is a re-scaling mask defined by a learnable mask $M$, which has the same shape as $\mathcal{D}(\textit{Conv}_{m\times m}(\widetilde{X}_{d}^{'}))$. The function $\phi(\cdot)$ represents the sigmoid activation, ensuring values remain within the range $[0,1]$.

Next, to enhance the representational capacity, we apply a linear expansion (Exp) that doubles the feature dimensions of $\text{SRF}$ results. Specifically, the first head is computed as:
\begin{equation} \label{eq_exp}
\text{Head}_{1}=\text{Exp}(\text{SRF}(\text{Conv}_{m\times m}(\widetilde{X}_{:d_{h}}^{'}))),        
\end{equation} 
where $\text{Head}_{1}\in \mathbb{R}^{2d_{h}\times H \times W}$ and $\widetilde{X}_{:d_{h}}^{'}\in \mathbb{R}^{d_h \times H \times W}$ represents the input for this head. The same operation is applied to the remaining heads.  Finally, all heads are concatenated along the feature dimension, followed by normalization and an activation function. A linear projection (Proj) then restores the feature dimensions to match $\widetilde{X}$, yielding the spectral-adaptive context aggregation $\mathcal{C}(\cdot)$:
\begin{equation} \label{eq_exp-proj}
\mathcal{C}(\widetilde{X})=\text{Proj}(\sigma(\text{Norm}([\text{Head}_{1};\text{Head}_{2};\cdots;\text{Head}_{N} ]))),              
\end{equation} 
where $[;;]$ denotes concatenation along the feature dimension, $\text{Norm}(\cdot)$ represents a normalization function, and $\mathcal{C}(\widetilde{X}) \in \mathbb{R}^{D\times H \times W}$ is the final aggregated output.

\subsection{The Spectral-Adaptive Modulation (SPAM) Mixer}
Building on ~\eqref{eq_conv_modulation}, we introduce SPAM, a novel token mixer that utilizes $\mathcal{V}(\cdot)$ and $\mathcal{C}(\cdot)$, as illustrated in Fig.~\ref{fig_SPAM}.

In SAG, leveraging the analysis in Fig.~\ref{fig_freq_response} to serve as a diverse spectral basis, we set $N=4$ and assign kernel sizes of $[3,5,7,9]$ to each head. GELU activation~\cite{hendrycks2016gaussian} is used for $\sigma(\cdot)$. To ensure stable training under our limited GPU resources, we apply a modified layer normalization for $\text{Norm}(\cdot)$, which can be implemented via PyTorch's GroupNorm API~\cite{paszke2019pytorch} with a group size of 1, normalizing the spatial dimension~\cite{yu2022metaformer}. Finally, after modulating $\mathcal{V}(\cdot)$ with $\mathcal{C}(\cdot)$, a linear transformation is applied to facilitate feature interaction, formulated as:
\begin{equation} \label{eq_SPAM}
\text{SPAM}(\widetilde{X})=\text{Linear}(\mathcal{V}(\widetilde{X}) \odot \mathcal{C}(\widetilde{X})).
\end{equation}


\subsection{SPANetV2 Architectures}
We introduce SPANetV2, a novel vision architecture that integrates the SPAM mixer within a four-stage hierarchical design, following the MetaFormer framework and its configurations~\cite{yu2023metaformer}. SPANetV2 consists of two model variants: pure models, which utilize a uniform token mixer, and hybrid models, which incorporate heterogeneous mixers.

\begin{table}
    \caption{Architecture configurations for SPANetV2.}
    \label{table:arch_config}
    \centering
    \resizebox{\columnwidth}{!}{%
    \begin{tabular}{c  c  c  c}
        \toprule
        Model size & Embedding dimensions  & Block numbers & Activation  \\ 
        \midrule
        S18 & [64, 128, 320, 512] & [3, 3, 9, 3]  &  GELU  \\  
        S36 & [64, 128, 320, 512] & [3, 12, 18, 3] & GELU  \\ 
        M36 & [96, 192, 384, 576] & [3, 12, 18, 3] & GELU  \\ 
        B36 & [128, 256, 512, 768] & [3, 12, 18, 3] & GELU \\       
        \bottomrule
    
    \end{tabular}%
    }
    \begin{tablenotes}[flushleft]
        \footnotesize
        \scriptsize
        \item[\hskip -\fontdimen 2 \font] Note: All models adopt the Metaformer framework~\cite{yu2023metaformer} for downsampling, using [7,3,3,3] kernels, [4,2,2,2] strides, and [2,1,1,1] padding at each stage.
    \end{tablenotes}
\end{table}

\subsubsection{Pure design}
Pure models follow all MetaFormer configurations, except replacing the token mixer with our SPAM and adopting GELU activation. Table~\ref{table:arch_config} details the embedding dimensions and block numbers for each model size.

\subsubsection{Hybrid design}
In hybrid settings, the first two stages employ SPAM, while the latter two stages incorporate heterogeneous mixers such as self-attention or else~\cite{yu2023metaformer,tatsunami2024fft}. To further enhance the spectral capabilities of self-attention, we integrate attention features with convolutional features through element-wise addition, defined as follows:
\begin{subequations}
\begin{align} \label{eq_mix_attn}
\textit{MixAttention}(X) &=  E(V) + \textit{Conv}_{m\times m}(\widetilde{V}), \\ 
V &= X^{T}W_{v}, 
\end{align}
\end{subequations}
where $\widetilde{V}\in \mathbb{R}^{D\times H \times W}$ is obtained by reshaping $V$ along the spatial dimensions, and $m=7$. The detailed application is shown in Table~\ref{table:Ablation}. This mixed attention improves convergence and enhances overall performance. In the following section, we evaluate SPANetV2 on ImageNet-1K classification—often associated with texture bias—and on several downstream tasks (\textit{e.g.}, object detection, instance segmentation, and semantic segmentation), where shape cues tend to be more critical~\cite{tuli2021convolutional, geirhos2018imagenet}.



\section{Experiments}
\label{sec:experiment}
Following established practices ~\cite{yu2023metaformer, li2023uniformer, rao2023gfnet, lin2014microsoft}, we conduct comprehensive experiments to evaluate the effectiveness of the proposed SPANetV2 across three tasks: image classification on ImageNet-1K ~\cite{deng2009imagenet}, object detection and instance segmentation on COCO ~\cite{lin2014microsoft}, and semantic segmentation on ADE20K ~\cite{zhou2017scene}. We first compare the performance of the proposed SPANetV2 with previous state-of-the-art methods across all three tasks. Additionally, we present an ablation study to assess the impact of individual design components within the proposed architecture. All experiments are implemented using PyTorch ~\cite{paszke2019pytorch} on an Ubuntu 22.04 system equipped with four NVIDIA RTX4090 GPUs.


\begin{table}
    \caption{Performance comparison of models trained on ImageNet-1k at the resolution of $224^{2}$. }
    \label{tab:imagenet-1k_results}
    \centering
    
    \resizebox{\columnwidth}{!}{
    \begin{tabular}{l c c | c c c}    
        \toprule
        Model & \makecell{Mixing \\ type} & \makecell{Params \\ (M)} & \makecell{MACs \\ (G)} & \makecell{Latency \\ (ms)} & \makecell{Top-1 \\ (\%)} \\ 
        \midrule

        RSB-ResNet-50~\cite{he2016deep, wightman2021resnet}   & C    & 26      & 4.1    & -    & 79.8    \\
        ConvNeXt-T~\cite{liu2022convnet}       & C    & 29      & 4.5    & 56.2    & 82.1      \\
        FocalNet-T (LRF)~\cite{yang2022focal}        & C    & 29      & 4.5    & 76.9    & 82.3      \\
        SPANet-S~\cite{yun2023spanet}          & C    & 29      & 4.6    & 127.6    & 83.1      \\
        SLaK-T~\cite{liu2023more}              & C    & 30      & 5.0    & -    & 82.5      \\
        InceptionNeXt-T~\cite{yu2024inceptionnext}        & C    & 28      & 4.2    & 53.3    & 82.3     \\
        Conv2Former-T~\cite{hou2024conv2former}              & C    & 27      & 4.4    & -    & 83.2      \\
        RDNet-T~\cite{kim2025densenets}        & C    & 24      & 5.1    & 70.9    & 82.8      \\
        ConvFormer-S18 \textdagger~\cite{yu2023metaformer} & C    & 27      & 3.9    & 84.0    & 83.0      \\
        \rowcolor{gray!10}
        \textbf{SPANetV2-S18-pure} (ours)      & C    & 29      & 4.2    & 103.0    & \textbf{83.4}       \\        
        \hdashline         
        DeiT-S~\cite{touvron2021training}      & A    & 22      & 4.6    & 31.0    & 79.8      \\
        Swin-T~\cite{liu2021swin}              & A    & 29      & 4.5    & 65.0    & 81.3      \\
        LITv2-S~\cite{pan2022hilo}             & A    & 28      & 3.7    &   -     & 82.0      \\        
        CSWin-T~\cite{dong2022cswin}           & A    & 23      & 4.3    & -     & 82.7      \\
        PoolFormerV2-S24~\cite{yu2023metaformer}   & P    & 21      & 3.4    & 86.7   & 80.7  \\
        DFFormer-S18 \textdagger~\cite{tatsunami2024fft}   & F    & 30      & 3.8    & 129.0   & 83.2      \\
        \hdashline         
        UniFormer-S~\cite{li2023uniformer}     & CA   & 22      & 3.6    & -     & 82.9      \\
        CAFormer-S18 \textdagger~\cite{yu2023metaformer} & CA    & 26      & 4.1    & 92.1   & 83.4      \\
        CDFFormer-S18 \textdagger~\cite{tatsunami2024fft}  & CF   & 30      & 3.9    & 105.3   & 83.1       \\
        \rowcolor{gray!10}
        \textbf{SPANetV2-S18-hybrid}  (ours)   & CA   & 27      & 4.2    & 104.9    & \textbf{83.9}       \\
        \midrule

        RSB-ResNet-101~\cite{he2016deep, wightman2021resnet}   & C    & 45      & 7.9    & -     & 81.3      \\
        ConvNeXt-S~\cite{liu2022convnet}       & C    & 50      & 8.7    & 89.5    & 83.1      \\
        FocalNet-S (LRF)~\cite{yang2022focal}        & C    & 50      & 8.7    & 120.5    & 83.5      \\
        SPANet-M~\cite{yun2023spanet}          & C    & 42      & 6.8    & 190.7    & 83.5      \\
        SLaK-S~\cite{liu2023more}              & C    & 55      & 9.8    & -    & 83.8      \\
        InceptionNeXt-S~\cite{yu2024inceptionnext}        & C    & 49      & 8.4    & 88.2    & 83.5      \\
        Conv2Former-S~\cite{hou2024conv2former}              & C    & 50      & 8.7    & -    & 84.1      \\
        RDNet-S~\cite{kim2025densenets}        & C    & 50      & 8.7    & 103.1    & 83.7      \\
        ConvFormer-S36 \textdagger~\cite{yu2023metaformer} & C    & 40      & 7.6   & 158.9    & 84.1      \\
        \rowcolor{gray!10}
        \textbf{SPANetV2-S36-pure}  (ours)     & C    & 44      & 8.1    & 198.4    & \textbf{84.4}       \\
        \hdashline 
        Swin-S~\cite{liu2021swin}              & A    & 50      & 8.7    & 104.4    & 83.0      \\
        LITv2-M~\cite{pan2022hilo}             & A    & 49      & 7.5    & -     & 83.3      \\
        CSWin-S~\cite{dong2022cswin}           & A    & 35      & 6.9    & -     & 83.6      \\
        PoolFormerV2-S36~\cite{yu2023metaformer}   & P    & 31      & 5.0    & 127.9   & 81.6  \\
        GFNet-H-S++~\cite{rao2023gfnet}        & F    & 37      & 4.6    & -     & 82.5      \\        
        DFFormer-S36 \textdagger~\cite{tatsunami2024fft}   & F    & 46      & 7.4   & 247.8    & 84.3      \\
        \hdashline 
        UniFormer-B~\cite{li2023uniformer}     & CA   & 50      & 8.3    & -     & 83.9      \\
        CAFormer-S36 \textdagger~\cite{yu2023metaformer}  & CA    & 39      & 8.0    & 174.9   & 84.4     \\
        CDFFormer-S36 \textdagger~\cite{tatsunami2024fft}  & CF   & 45      & 7.5   & 200.0    & 84.2     \\
        \rowcolor{gray!10}
        \textbf{SPANetV2-S36-hybrid}  (ours)   & CA   & 41      & 8.1    & 203.1    & \textbf{84.7}       \\
        \midrule

        RSB-ResNet-152~\cite{he2016deep, wightman2021resnet}   & C    & 60      & 11.6    & -     & 81.8      \\
        RepLKNet-31B~\cite{ding2022scaling}     & C    & 79     & 15.3    & -     & 83.5        \\
        ConvNeXt-B~\cite{liu2022convnet}        & C    & 89     & 15.4    & 133.3    & 83.8          \\
        FocalNet-B (LRF)~\cite{yang2022focal}         & C    & 89     &15.4    & 179.5    & 83.9      \\
        SPANet-B~\cite{yun2023spanet}           & C    & 76     & 12.0    & 293.2    & 84.0      \\
        SLaK-B~\cite{liu2023more}              & C    & 95      & 15.4    & -    & 84.0      \\
        InceptionNeXt-B~\cite{yu2024inceptionnext}        & C    & 87      & 14.9    & 133.0    & 84.0      \\        
        Conv2Former-B~\cite{hou2024conv2former}              & C    & 90      & 15.9    & -    & 84.4      \\
        RDNet-B~\cite{kim2025densenets}        & C    & 88      & 15.4    & 154.0   & 84.4      \\
        ConvFormer-M36 \textdagger~\cite{yu2023metaformer} & C    & 57      & 12.8   & 224.1   & 84.5      \\
        \rowcolor{gray!10}
        \textbf{SPANetV2-M36-pure}  (ours)      & C    & 61      & 13.7   & 276.4    & \textbf{84.9}      \\
        \hdashline 
        DeiT-B~\cite{touvron2021training}       & A    & 86     & 17.5   & 86.4    & 81.8      \\
        Swin-B~\cite{liu2021swin}               & A    & 88      & 15.4    & 153.4    & 83.5        \\
        LITv2-B~\cite{pan2022hilo}              & A    & 87     &13.2    & -     & 83.6          \\
        CSWin-B~\cite{dong2022cswin}            & A    & 78     & 15.0   & -     & 84.2        \\
        MaxViT-S~\cite{tu2022maxvit}           & A    & 69      & 11.7   & 223.6    & 84.5      \\
        PoolFormerV2-M36~\cite{yu2023metaformer}   & P    & 56      & 8.8    & 191.0   & 82.2  \\
        GFNet-H-B++~\cite{rao2023gfnet}        & F    & 62      & 8.6    & -    & 83.5      \\
        DFFormer-M36 \textdagger~\cite{tatsunami2024fft}   & F    & 65      & 12.5   & 358.7    & 84.6      \\
        \hdashline 
        CAFormer-M36 \textdagger~\cite{yu2023metaformer} & CA    & 56      & 13.2    & 242.9   & 85.1     \\
        CDFFormer-M36 \textdagger~\cite{tatsunami2024fft}  & CF   & 64      & 12.7   & 287.0    & 84.8     \\
        \rowcolor{gray!10}
        \textbf{SPANetV2-M36-hybrid}  (ours)    & CA   & 58      & 13.6   & 278.8    & \textbf{85.3}      \\
        \midrule

        ConvNeXt-L~\cite{liu2022convnet}        & C    & 198    & 34.4   & 237.1    & 84.3        \\
        RDNet-L~\cite{kim2025densenets}         & C    & 186    & 34.7   & 266.1    & 84.8        \\
        ConvFormer-B36~\cite{yu2023metaformer}  & C    & 100    & 22.6   & 314.6   & 84.8        \\
        \rowcolor{gray!10}
        \textbf{SPANetV2-B36-pure}  (ours)      & C    & 100    & 24.3   & 395.4    & \textbf{85.0}     \\
        \hdashline 
        MaxViT-B~\cite{tu2022maxvit}            & A    & 120    & 23.4   & 388.9    & 85.0        \\
        PoolFormerV2-M48~\cite{yu2023metaformer}   & P    & 73      & 11.5    & 252.3    & 82.6      \\
        GFNet-H-L++~\cite{rao2023gfnet}         & F    & 110    & 15.3   & -    & 83.9         \\
        DFFormer-B36~\cite{tatsunami2024fft}    & F    & 115    & 22.1   & 500.0    & 84.8        \\
        \hdashline 
        CAFormer-B36~\cite{yu2023metaformer}    & CA    & 99      & 23.2    & 339.3    & 85.5    \\
        CDFFormer-B36~\cite{tatsunami2024fft}   & CF   & 113    & 22.5   & 405.3   & 85.0    \\
        \rowcolor{gray!10}
        \textbf{SPANetV2-B36-hybrid}  (ours)    & CA   & 100    & 23.9   & 398.5    & \textbf{85.6}    \\
        \bottomrule 
       
    \end{tabular}
    }
    
    \begin{tablenotes}[flushleft]
        \footnotesize
        \scriptsize
        \item[\hskip -\fontdimen 2 \font] Note: The symbol '\textdagger' indicates models retrained from scratch. The full names of the mixing types are as follows. C denotes Convolution, A denotes Attention, P denotes Pooling, F denotes FFT, CA represents a hybrid of Convolution and Attention, and CF represents a hybrid of Convolution and FFT. The inference time (ms) is measured for $128$ images of size $224\times224$ on an RTX4090 GPU.
    \end{tablenotes}

\end{table}

\subsection{Image Classification on ImageNet-1K}

\subsubsection{Setup}
ImageNet-1K~\cite{deng2009imagenet} is a widely cited benchmark in computer vision, comprising $1.28M$ training and $50K$ validation images across $1K$ classes. Following the MetaFormer baseline settings~\cite{yu2023metaformer}, all models are trained for 300 epochs at $224^2$ resolution. Data augmentation and regularization include MixUp~\cite{zhang2018mixup}, CutMix~\cite{yun2019cutmix}, CutOut~\cite{zhong2020random}, RandAugment~\cite{cubuk2020randaugment}, Label Smoothing~\cite{szegedy2016rethinking}, and Stochastic Depth~\cite{huang2016deep}. We do not use repeated augmentation ~\cite{berman2019multigrain,hoffer2020augment} and LayerScale~\cite{touvron2021going}, but ResScale~\cite{shleifer2021normformer} is applied in the last two stages to facilitate deep model training. We adopt AdamW~\cite{kingma2014adam, loshchilov2017decoupled} optimizer with weight decay $0.05$ and a peak learning rate of $\text{lr} = 1e^{-3} \times \frac{\text{batch size}}{1024}$ for pure models. In contrast, LAMB~\cite{you2019large} optimizer is adopted and a peak learning rate is set to $\text{lr} = 2e^{-3} \times \frac{\text{batch size}}{1024}$ for hybrid models. All models are trained with a batch size of 4096. The number of warmup epochs is 5, and cosine schedule is used to decay the learning rate. Our implementation is based on PyTorch-image-models~\cite{rw2019timm} and the MetaFormer baseline~\cite{yu2023metaformer}.

\subsubsection{Results} 
Table~\ref{tab:imagenet-1k_results} presents the ImageNet-1K classification performance of SPANetV2. Our models achieve superior top-1 accuracy across S18, S36, M36, and B36 configurations in both pure and hybrid settings, surpassing various token-mixing approaches. Specifically, SPANetV2-S18-pure outperforms other convolution-based networks, such as multi-size kernel convolutions like FocalNet-T ($1.1\%{p}$) and InceptionNeXt-T ($1.4\%{p}$), and large-kernel designs including SLaK-T ($0.9\%{p}$) and Conv2Former-T ($0.2\%{p}$). Furthermore, it surpasses LITv2-S ($1.4\%{p}$), which employs an attention-based mixer for low- and high-frequency processing, and DFFormer-S18 ($0.2\%{p}$), which utilizes an input-adaptive dynamic spectral filter. Finally, SPANetV2-S18-hybrid surpasses all models, outperforming even other hybrid variants, exceeding CAFormer-S18 ($0.5\%{p}$) and CDFFormer-S18 ($0.8\%{p}$). Similar results are also observed for S36, M36, and B36.

\subsection{Object Detection and Instance Segmentation on COCO}

\subsubsection{Setup}
We evaluate SPANetV2 on the COCO benchmark~\cite{lin2014microsoft}, which comprises $118K$ training (\texttt{train2017}) and $5K$ validation (\texttt{val2017}) images. SPANetV2 serves as the backbone for Mask R-CNN~\cite{he2017mask} and Cascade Mask R-CNN~\cite{cai2018cascade}. Backbones are initialized with ImageNet pre-trained weights, while added layers use Xavier initialization~\cite{glorot2010understanding}. Following ConvNeXt-based settings~\cite{liu2022convnet}, models are trained for 36 epochs ($3\times$ schedule) with a batch size of 16. AdamW~\cite{kingma2014adam, loshchilov2017decoupled} is used with an initial learning rate of $1e^{-4}$ for Mask R-CNN and $2e^{-4}$ for Cascade Mask R-CNN. Training and testing images are resized to a shorter side of 800 pixels, with a maximum longer side of $1,333$ pixels. Implementation is based on the $\texttt{mmdetection}$~\cite{chen2019mmdetection} codebase. 

\subsubsection{Results} As shown in Table~\ref{table:coco}, SPANetV2 models with Mask R-CNN~\cite{he2017mask} outperform their counterparts. SPANetV2-S18-pure achieves $48.0 \ \textit{AP}^{b}$ and $42.9 \ \textit{AP}^{m}$, surpassing Swin-T ($46.0 \ \textit{AP}^{b}$, $41.6 \ \textit{AP}^{m}$), ConvNeXt-T ($46.2 \ \textit{AP}^{b}$, $41.7 \ \textit{AP}^{m}$), RDNet-T ($47.5 \ \textit{AP}^{b}$, $42.4 \ \textit{AP}^{m}$), and ConvFormer-S18 ($47.7 \ \textit{AP}^{b}$, $42.6 \ \textit{AP}^{m}$), while achieving comparable performance to Conv2Former-T ($48.0 \ \textit{AP}^{b}$, $43.0 \ \textit{AP}^{m}$). Meanwhile, SPANetV2-S18-hybrid outperforms all models, attaining $49.6 \ \textit{AP}^{b}$ and $44.3 \ \textit{AP}^{m}$, exceeding the hybrid model CAFormer-S18 ($48.6 \ \textit{AP}^{b}$, $43.7 \ \textit{AP}^{m}$). Furthermore, SPANetV2-S18-pure with Cascade Mask R-CNN~\cite{cai2018cascade} surpasses all pure models, while SPANetV2-S18-hybrid outperforms all models, including CAFormer-S18. Due to our resource constraints, only S18 models are evaluated.

\begin{table}
    \caption{Performance of object detection with Mask R-CNN and Cascade Mask R-CNN on COCO val2017.} 
    \label{table:coco}
    \centering
    
    \resizebox{\columnwidth}{!}{
    \begin{tabular}{lcccccccc}
        \toprule
        
        Backbone & MACs (G) & FPS & $AP^{b}$ & $AP^{b}_{50}$ & $AP^{b}_{75}$ & $AP^{m}$  & $AP^{m}_{50}$ & $AP^{m}_{75}$ \\    
        \midrule

        \multicolumn{1}{l}{} & \multicolumn{8}{c}{Mask R-CNN $3\times$} \\ 
        Swin-T~\cite{liu2021swin}               & 267 & 27.4 & 45.7 & 68.1 & 50.3 & 41.6 & 65.1 & 44.9  \\
        ConvNeXt-T~\cite{liu2022convnet}        & 262 & 27.5 & 45.9 & 67.9 & 50.8 & 41.7 & 65.0 & 44.9  \\
        Conv2Former-T~\cite{hou2024conv2former} & -   & - & 48.0 & 69.5 & 52.7 & 43.0 & 66.8 & 46.1  \\   
        RDNet-T~\cite{kim2025densenets}         & 278 & 14.1 & 47.5 & 68.5 & 52.1 & 42.4 & 65.6 & 45.7  \\        
        ConvFormer-S18~\cite{yu2023metaformer}  & 251 & 26.2 & 47.7 & 69.6 & 52.3 & 42.6 & 66.3 & 45.9  \\
        \rowcolor{gray!10} \textbf{SPANetV2-S18-pure} (ours) & 255 & 10.5 & \textbf{48.0} & 69.7 & 52.8 & \textbf{42.9} & 66.7 & 46.2  \\
        CAFormer-S18~\cite{yu2023metaformer}    & 254 & 12.8 & 48.6 & 70.5 & 53.4 & 43.7 & 67.5 & 47.4  \\
        \rowcolor{gray!10} \textbf{SPANetV2-S18-hybrid} (ours) & 251 & 9.2 & \textbf{49.6} & 71.3 & 54.8 & \textbf{44.3} & 68.3 & 48.1  \\        
        \midrule

        \multicolumn{1}{l}{} & \multicolumn{8}{c}{Cascade Mask R-CNN $3\times$} \\ 
        Swin-T~\cite{liu2021swin}               & 745 & 16.0 & 50.4 & 69.2 & 54.7 & 43.7 & 66.6 & 47.3  \\
        ConvNeXt-T~\cite{liu2022convnet}        & 741 & 16.7 & 50.4 & 69.1 & 54.8 & 43.7 & 66.5 & 47.3  \\
        SLaK-T~\cite{liu2023more}               & - & - & 51.3 & 70.0 & 55.7 & 44.3 & 67.2 & 48.1  \\   
        Conv2Former-T~\cite{hou2024conv2former} & - & - & 51.4 & 69.8 & 55.9 & 44.5 & 67.4 & 48.3  \\   
        ConvFormer-S18~\cite{yu2023metaformer} & 729 & 16.3 & 51.5 & 70.7 & 55.8 & 44.6 & 67.8 & 48.2  \\
        \rowcolor{gray!10} \textbf{SPANetV2-S18-pure} (ours) & 734 & 8.6  & \textbf{51.6} & 70.3 & 55.9 & \textbf{44.7} & 68.0 & 48.1  \\
        CAFormer-S18~\cite{yu2023metaformer}   & 733 & 9.8 & 52.3 & 71.3 & 56.9 & 45.2 & 68.6 & 48.8  \\
        \rowcolor{gray!10} \textbf{SPANetV2-S18-hybrid} (ours) & 729 & 7.7 & \textbf{52.8} & 71.4 & 57.5 & \textbf{45.7} & 68.9 & 49.5  \\
        \hdashline   
        Swin-S~\cite{liu2021swin}               & 838 & 13.0 & 51.9 & 70.7 & 56.3 & 45.0 & 68.2 & 48.8  \\
        ConvNeXt-S~\cite{liu2022convnet}        & 827 & 13.8 & 51.9 & 70.8 & 56.5 & 45.0 & 68.4 & 49.1  \\        
        Conv2Former-S~\cite{hou2024conv2former} & - & - & 52.8 & 71.4 & 57.3 & 45.7 & 69.0 & 49.8  \\   
        ConvFormer-S36~\cite{yu2023metaformer} & 805 & 14.4 & 52.5 & 71.1 & 57.0 & 45.2 & 68.6 & 48.8  \\
        \rowcolor{gray!10} \textbf{SPANetV2-S36-pure} (ours) & 814 & 5.6  & \textbf{53.1} & 72.2 & 57.6 & \textbf{45.9} & 69.8 & 49.4  \\
        CAFormer-S36~\cite{yu2023metaformer}   & 811 & 6.6 & 53.2 & 72.1 & 57.7 & 46.0 & 69.5 & 49.8  \\
        \rowcolor{gray!10} \textbf{SPANetV2-36-hybrid} (ours) & 806 & 5.1 & \textbf{53.8} & 72.9 & 58.4 & \textbf{46.4} & 70.0 & 50.2  \\        
        \bottomrule
    
    \end{tabular}
    }

    \begin{tablenotes}[flushleft]
        \footnotesize
        \item[\hskip -\fontdimen 2 \font] Note: The MACs are measured at a resolution of $1280\times800$. Performance is reported in terms of bounding box $\text{AP}$ and mask $\text{AP}$, denoted by $\text{AP}^{b}$ and $\text{AP}^{m}$, respectively. The FPS are measured on an A100 GPU.
    \end{tablenotes}

\end{table}

\subsection{Semantic Segmentation on ADE20K}

\subsubsection{Setup} We also evaluate SPANetV2 on the ADE20K~\cite{zhou2017scene} for semantic segmentation, using $20K$ training and $2K$ validation images across $150$ categories. We employ UperNet~\cite{xiao2018unified} and Semantic FPN~\cite{kirillov2019panoptic} as semantic segmentation models, using $\texttt{mmsegmentation}$\cite{mmseg2020} as the base framework with SPANetV2 as the backbone. Backbones are initialized with ImageNet-1K pre-trained weights, while newly added layers are initialized using Xavier initialization\cite{glorot2010understanding}. For UperNet training, we follow the ConvNeXt-based settings~\cite{liu2022convnet}, training for $160K$ iterations with a batch size of $16$, using the AdamW optimizer~\cite{kingma2014adam, loshchilov2017decoupled} with an initial learning rate of $1e^{-4}$. For Semantic FPN, we adopt the PoolFormer-based settings~\cite{yu2022metaformer}, training for $40K$ iterations with a batch size of $32$, using AdamW with an initial learning rate of $2e^{-4}$. Training images are randomly resized and cropped to $512\times512$, while testing images are rescaled to a shorter side of $512$ pixels.

\subsubsection{Results}
As shown in Table~\ref{table:ADE20K}, SPANetV2 with UperNet~\cite{xiao2018unified} outperforms existing models. Specifically, SPANetV2-S18-pure achieves $48.7\%$ mIoU, surpassing all pure models, while SPANetV2-S18-hybrid attains $49.1\%$, outperforming all models, including the hybrid CAFormer-S18. For S36 models, SPANetV2-S36-pure achieves $49.8\%$ mIoU with a marginal improvement, whereas SPANetV2-S36-hybrid sets a new state-of-the-art by reaching $51.2\%$ mIoU. Similarly, SPANetV2 integrated with Semantic FPN~\cite{kirillov2019panoptic} consistently outperforms prior models in semantic segmentation. SPANetV2-S18-pure achieves $46.7\%$ mIoU, surpassing all pure models, while SPANetV2-S18-hybrid attains $47.8\%$, outperforming all models, including the hybrid CDFFormer-S18. Comparable trends are also observed in S36 models.

\begin{table}    
    \caption{Performance of semantic segmentation with UperNet and Semantic FPN on ADE20K.}
    \label{table:ADE20K}
    \centering
    
    \resizebox{\columnwidth}{!}{
    \begin{tabular}{lcccc}
        \toprule
        Backbone & Params (M) & MACs (G) & FPS & mIoU (\%) \\ 
        \midrule
        \multicolumn{1}{l}{} & \multicolumn{4}{c}{UperNet} \\ 
        Swin-T~\cite{liu2021swin}                   & 60      & 945  & 40.8 & 45.8 \\
        ConvNeXt-T~\cite{liu2022convnet}             & 60      & 939   & 36.6 & 46.7 \\
        Conv2Former-T~\cite{hou2024conv2former}       & 56      & -   & - & 48.0 \\
        ConvFormer-S18~\cite{yu2023metaformer}       & 54      & 925   & 36.6 & 48.6 \\
        \rowcolor{gray!10} \textbf{SPANetV2-S18-pure} (ours)   & 60    & 930 & 18.8  & \textbf{48.7} \\ 
        CAFormer-S18~\cite{yu2023metaformer}         & 54      & 1024  & 33.0 & 48.9 \\
        \rowcolor{gray!10} \textbf{SPANetV2-S18-hybrid} (ours) & 58  & 925  & 21.9 & \textbf{49.1} \\
        \hdashline         
        Swin-S~\cite{liu2021swin}                   & 81      & 1038  & 28.7 & 49.5 \\
        ConvNeXt-S~\cite{liu2022convnet}             & 82      & 1027  & 27.7 & 49.6 \\
        Conv2Former-S~\cite{hou2024conv2former}       & 79      & -   & - & 50.3 \\
        ConvFormer-S36~\cite{yu2023metaformer}       & 67      & 1003  & 23.0 & 50.7 \\
        \rowcolor{gray!10} \textbf{SPANetV2-S36-pure} (ours)   & 75     & 1012  & 10.8 & \textbf{49.8} \\ 
        CAFormer-S36~\cite{yu2023metaformer}         & 67      & 1197  & 20.6 & 50.8 \\
        \rowcolor{gray!10} \textbf{SPANetV2-S36-hybrid} (ours) & 71  & 1004  & 11.9 & \textbf{51.6} \\        
        
        \midrule
        \multicolumn{1}{l}{} & \multicolumn{4}{c}{Semantic FPN} \\ 
        ResNet-50~\cite{he2016deep}                  & 29  & 46   & 33.5 & 36.7 \\
        Swin-T~\cite{liu2021swin}                    & 32  & 46   & 47.8 & 41.5 \\
        PoolFormer-S24~\cite{yu2022metaformer}       & 25  & 39   & 34.2 & 40.3 \\
        LITv2-S~\cite{pan2022hilo}                   & 31  & 41   & - & 44.3 \\                
        SPANet-S~\cite{yun2023spanet}                & 32  & 46   & 10.6 & 45.4 \\
        InceptionNeXt-T~\cite{yu2024inceptionnext}   & 28  & 44   & - & 43.1 \\        
        DFFormer-S18~\cite{tatsunami2024fft}         & 35  & 41   & 17.0 & 45.1 \\
        \rowcolor{gray!10} \textbf{SPANetV2-S18-pure} (ours)   & 33  & 43  & 18.0 & \textbf{46.7} \\ 
        CDFFormer-S18~\cite{tatsunami2024fft}        & 35  & 42  & 26.3 & 44.9 \\
        \rowcolor{gray!10} \textbf{SPANetV2-S18-hybrid} (ours) & 31  & 48  & 24.7 & \textbf{47.8} \\ 
        \hdashline         
    
        ResNet-101~\cite{he2016deep}                  & 48  & 65   & 33.5 & 38.8 \\
        Swin-S~\cite{liu2021swin}                    & 53  & 70   & 31.1 & 45.2 \\
        PoolFormer-S36~\cite{yu2022metaformer}       & 35  & 48    & 25.7 & 42.0 \\
        LITv2-M~\cite{pan2022hilo}                   & 52  & 63   & - & 45.7 \\
        SPANet-M~\cite{yun2023spanet}                & 45  & 57   & 8.2 & 46.2 \\   
        InceptionNeXt-S~\cite{yu2024inceptionnext}   & 50  & 65   & - & 45.6 \\
        DFFormer-S36~\cite{tatsunami2024fft}         & 50  & 60    & 10.5 & 47.5 \\
        \rowcolor{gray!10} \textbf{SPANetV2-S36-pure} (ours)   & 48  & 64  & 11.4 & \textbf{47.9} \\ 
        CDFFormer-S36~\cite{tatsunami2024fft}        & 50  & 61    & 19.8 & 46.7 \\
        \rowcolor{gray!10} \textbf{SPANetV2-S36-hybrid} (ours) & 45  & 74  & 14.3 & \textbf{48.6} \\ 
        \bottomrule
    \end{tabular}
    }
    
    \begin{tablenotes}[flushleft]
        \footnotesize
        \item[\hskip -\fontdimen 2 \font] Note: The MACs are measured at a resolution of $512\times 2048$ for UperNet and $512\times 512$ for Semantic FPN, respectively. The FPS are measured on an A100 GPU.
    \end{tablenotes}

\end{table}

\subsection{Ablation}
This section presents ablation studies on SPANetV2-S18 using ImageNet-1K~\cite{deng2009imagenet}. The results, summarized in Table~\ref{table:Ablation}, are analyzed below across key aspects.

\subsubsection{SPAM components} 
To assess the significance of SPAM components, we conduct experiments by modifying its operators. First, SRF is identified as a critical element of SAG, as its removal leads to a performance drop of $-0.2\%{p}$. Next, we configure SAG with a single SRF shared across all features in each head, achieving results identical to depth-wise SRF applied per feature dimension. However, using a single SRF reduces training stability as model size increases. Consequently, we employ depth-wise SRF for all SPANetV2 variants. Finally, we investigate the impact of multi-scale kernels by replacing them with single-scale counterparts. This modification inherently limits the spectral modulation to a narrow, fixed frequency range determined by the kernel size. As shown in Table~\ref{table:Ablation}, this restriction negatively impacts performance, resulting in accuracy drops for $3\times3$ ($-0.3\%_{p}$), $5\times5$ ($-0.3\%_{p}$), and $9\times9$ ($-0.2\%_{p}$) kernels. These results corroborate that aggregating diverse spectral components across multiple scales is a crucial factor for improving the performance of vision models.

\subsubsection{Hybrid stages}
We evaluate SPAM-based token mixers by comparing different stacking strategies, including SepConv, Attention, and MixAttention. As shown in Table~\ref{table:Ablation}, a uniform stacking of SPAM yields superior results compared to using only SepConv, highlighting the significance of spectral-adaptive convolution in enhancing model performance. Incorporating SPAM in the bottom two stages further improves performance over the SepConv-only configuration.

Next, we assess the impact of Attention and MixAttention in the bottom stages. Utilizing only Attention, as in CAFormer~\cite{yu2023metaformer}, improves SPANetV2-S18 performance by $0.3\%{p}$, outperforming the configuration that solely employs MixAttention. However, combining MixAttention in the third stage with Attention in the final stage yields the highest improvement of $0.5\%{p}$. Based on these findings, we adopt this configuration for all hybrid models.

\subsubsection{Branch output scaling} 
Branch output scaling is evaluated based on the MetaFormer block, revealing that ResScale~\cite{shleifer2021normformer} is the most effective for SPANetV2. Notably, employing LayerScale~\cite{touvron2021going} or combining both techniques ~\cite{yu2023metaformer} leads to a performance drop of $-0.1\%{p}$, while omitting scaling entirely further degrades performance by $-0.2\%{p}$.

\subsubsection{Biases in each block}
Enabling biases in each block results in a performance drop of $-0.1\%_{p}$. This finding aligns with Yu \textit{et al.}~\cite{yu2023metaformer}. Consequently, we disable biases by default for all SPANetV2 models, following prior studies\cite{raffel2020exploring,chowdhery2023palm}.


\begin{table}
    \caption{Ablation for SPANetV2-S18 on ImageNet-1K.}
    \label{table:Ablation}
    \centering
    \resizebox{\columnwidth}{!}{%
    \begin{tabular}{l | l | c}
        \toprule
        Ablation &  Variant  &  Top-1 (\%) \\ 
        \midrule
        - & SPANetV2-S18-pure & 83.4 \\ 
        \midrule  

        \multirow{6}{*}{SPAM components}
        & SAG with depth-wise SRF          &       \\ 
        & → without SRF & 83.2 (\textcolor{red}{-0.2}) \\
        & → with a single SRF & 83.4 (+0.0) \\  
        & SAG with multi-scale kernels           &       \\ 
        & → Single-scale $3\times3$          &  83.1 (\textcolor{red}{-0.3})    \\ 
        & → Single-scale $5\times5$          &  83.1 (\textcolor{red}{-0.3})    \\ 
        & → Single-scale $9\times9$          &  83.2 (\textcolor{red}{-0.2})    \\ 
      
        \midrule
    
        \multirow{6}{*}{Hybrid stages}
        & {[}SPAM, SPAM, SPAM, SPAM{]}   &                              \\ 
        & → {[}SepConv, SepConv, SepConv, SepConv{]} & 83.0 (\textcolor{red}{-0.4}) \\ 
        & → {[}SepConv, SepConv, SPAM, SPAM{]} & 83.1 (\textcolor{red}{-0.3}) \\ 
        & → {[}SPAM, SPAM, Attention, Attention{]} & 83.7 (\textcolor{blue}{+0.3}) \\ 
        & → {[}SPAM, SPAM, MixAttention, MixAttention{]} & 83.6 (\textcolor{blue}{+0.2}) \\ 
        & → {[}SPAM, SPAM, MixAttention, Attention{]} & 83.9 (\textcolor{blue}{+0.5}) \\         
        \midrule
    
        \multirow{4}{*}{Branch output scaling} 
        & ResScale~\cite{shleifer2021normformer}  &  \\ 
        & → None                                  & 83.2 (\textcolor{red}{-0.2}) \\ 
        & → LayerScale~\cite{touvron2021going}    & 83.3 (\textcolor{red}{-0.1}) \\ 
        & → BranchScale~\cite{yu2023metaformer}   & 83.3 (\textcolor{red}{-0.1}) \\ 
        \midrule
    
        \multirow{2}{*}{Biases in each block}
        & Disable biases of Norm, FC and Conv &                              \\ 
        & → Enable biases                     & 83.3 (\textcolor{red}{-0.1}) \\     
        \bottomrule

    \end{tabular}%
    }

    \begin{tablenotes}[flushleft]
        \footnotesize
        \item[\hskip -\fontdimen 2 \font] Note: SepConv is employed in ConvFormer-S18~\cite{yu2023metaformer}.
    \end{tablenotes}

\end{table}

\section{Comparison Analysis}
\label{sec:analysis}
We analyze the feature representations of SPANetV2 and other MetaFormer-based models listed in Table~\ref{tab:imagenet-1k_results}, including ConvFormer, CAFormer~\cite{yu2023metaformer}, DFFormer, and CDFFormer~\cite{tatsunami2024fft}. While maintaining identical architectural configurations, these models only differ in their token mixers, such as SepConv, Attention, and Dynamic Filter, allowing for a fair comparison of their spectral filtering capabilities.

\subsection{Relative Log Amplitude Analysis}
\label{sec:log-amp}
\begin{figure}[htbp]
\centering
\subfloat{\includegraphics[width=0.234\textwidth]{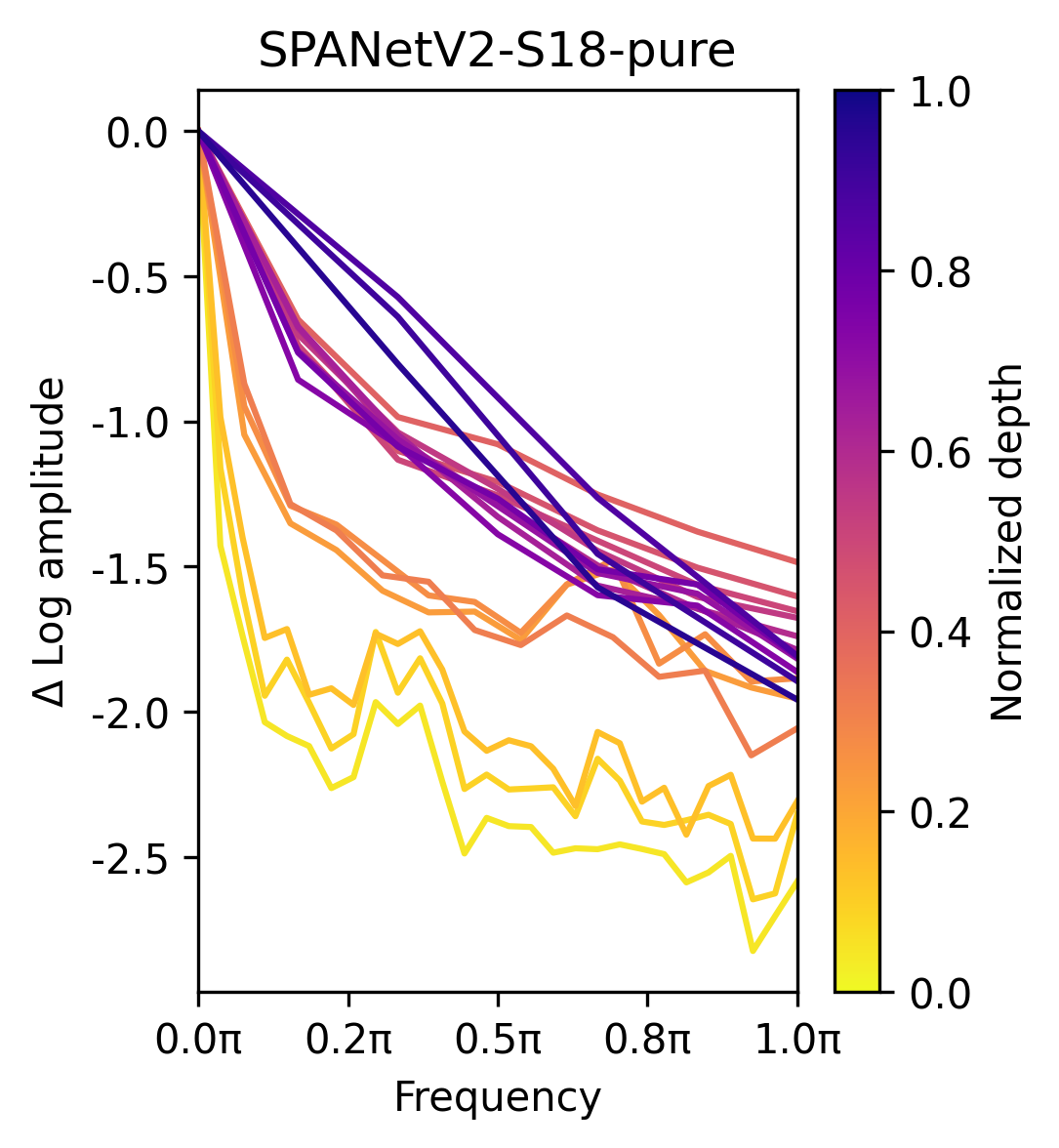}} \quad
\subfloat{\includegraphics[width=0.234\textwidth]{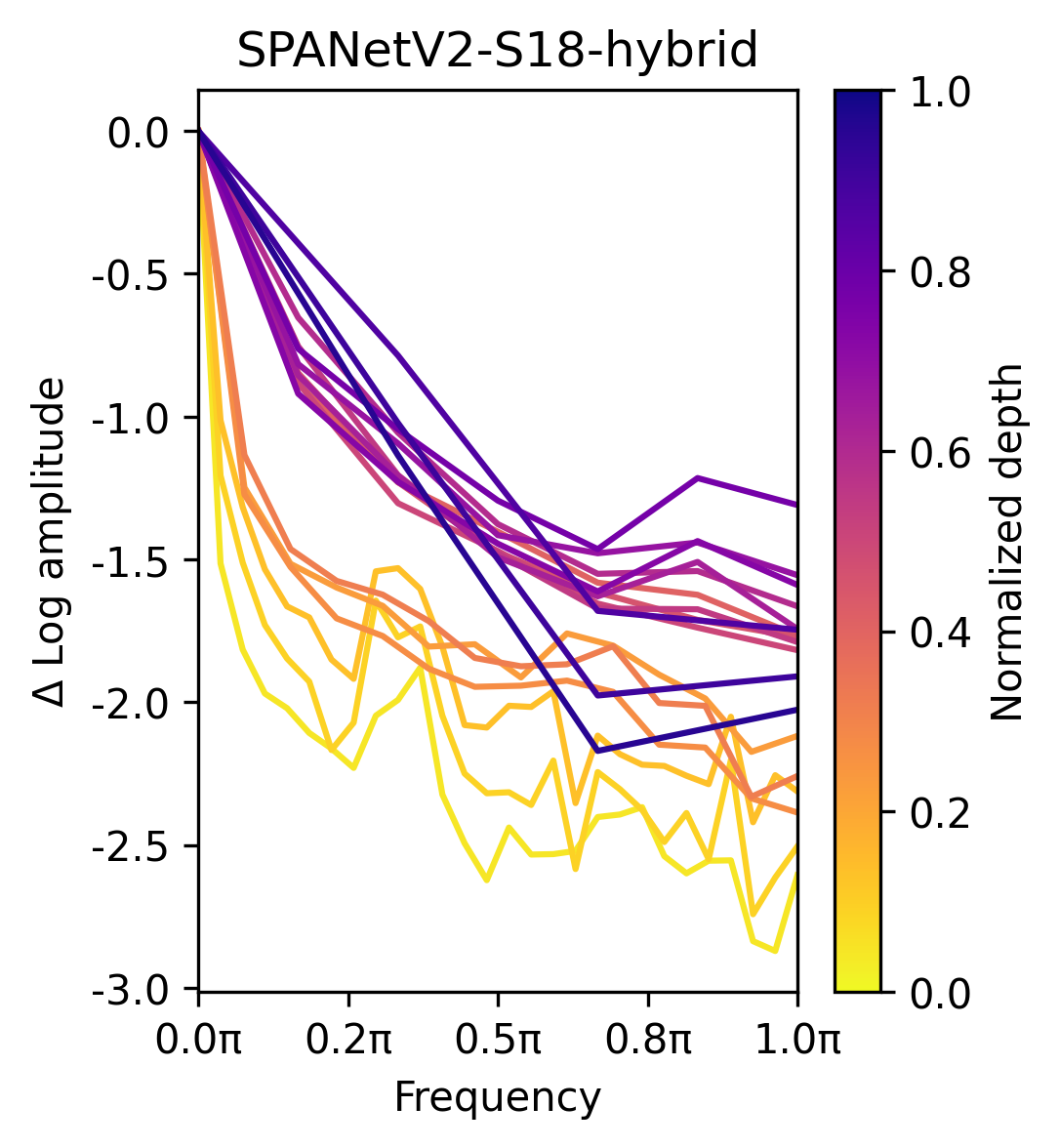}} \\  [-1em]   

\subfloat{\includegraphics[width=0.234\textwidth]{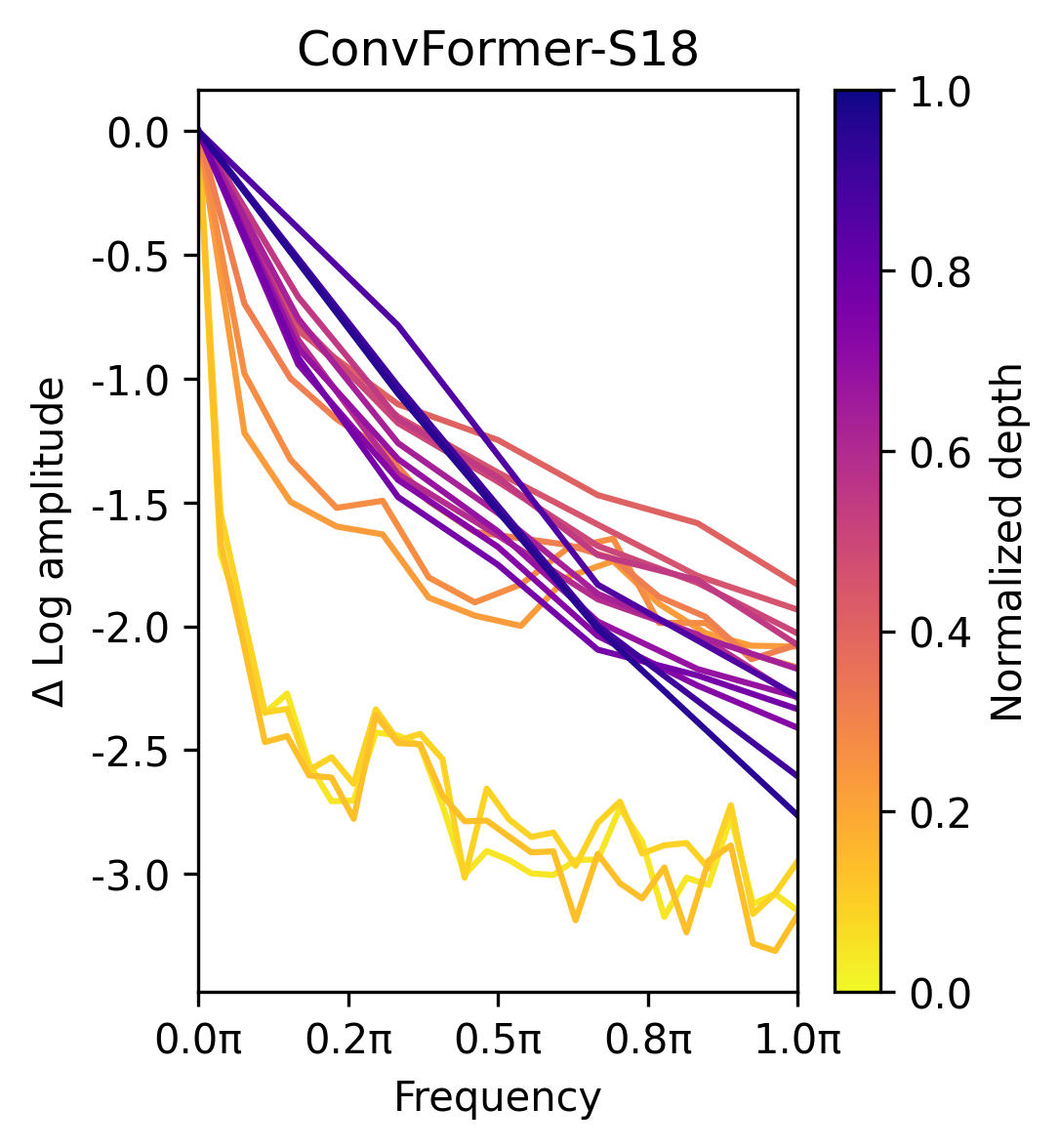}} \quad
\subfloat{\includegraphics[width=0.234\textwidth]{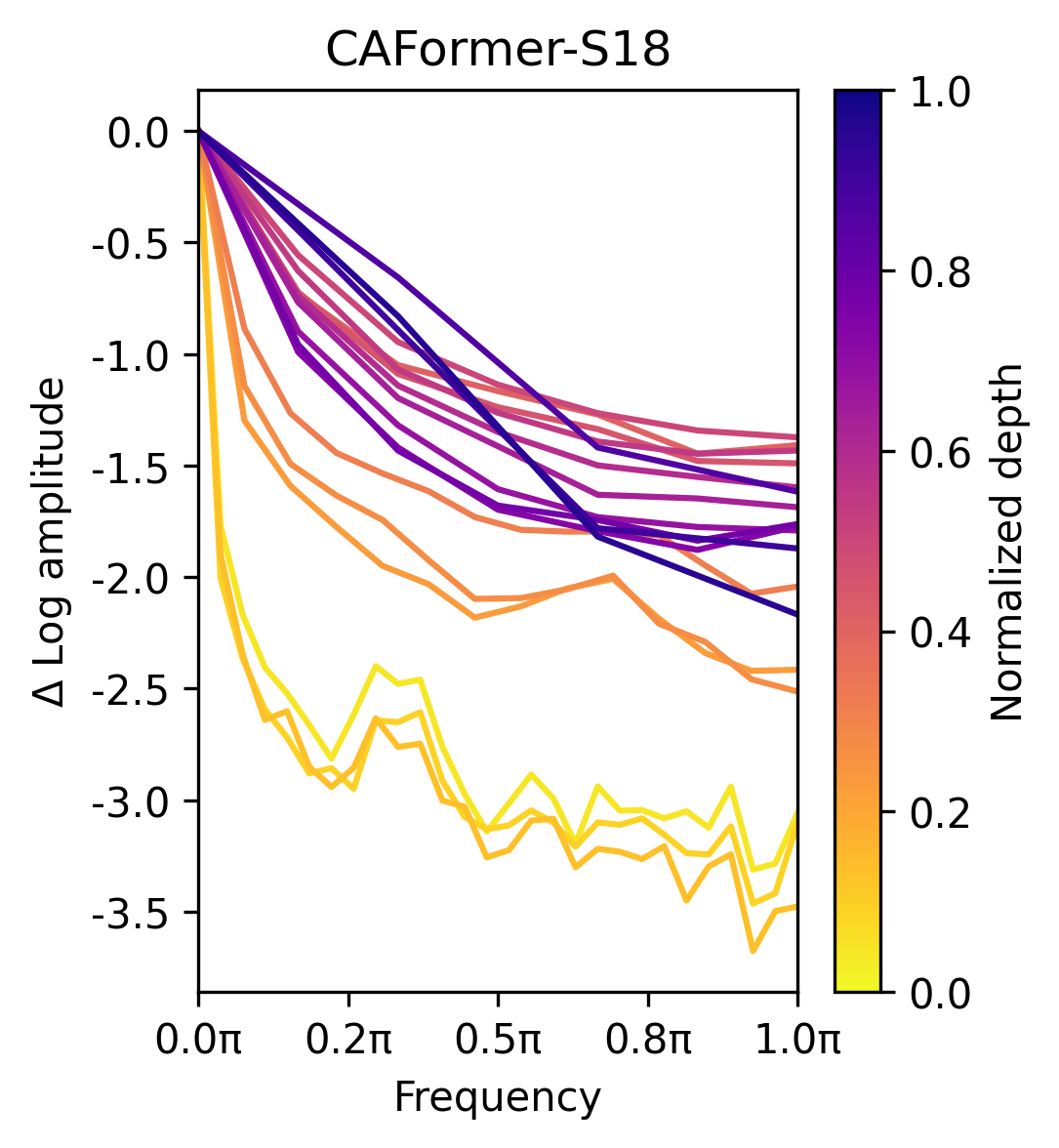}} \\  [-1em]  

\subfloat{\includegraphics[width=0.234\textwidth]{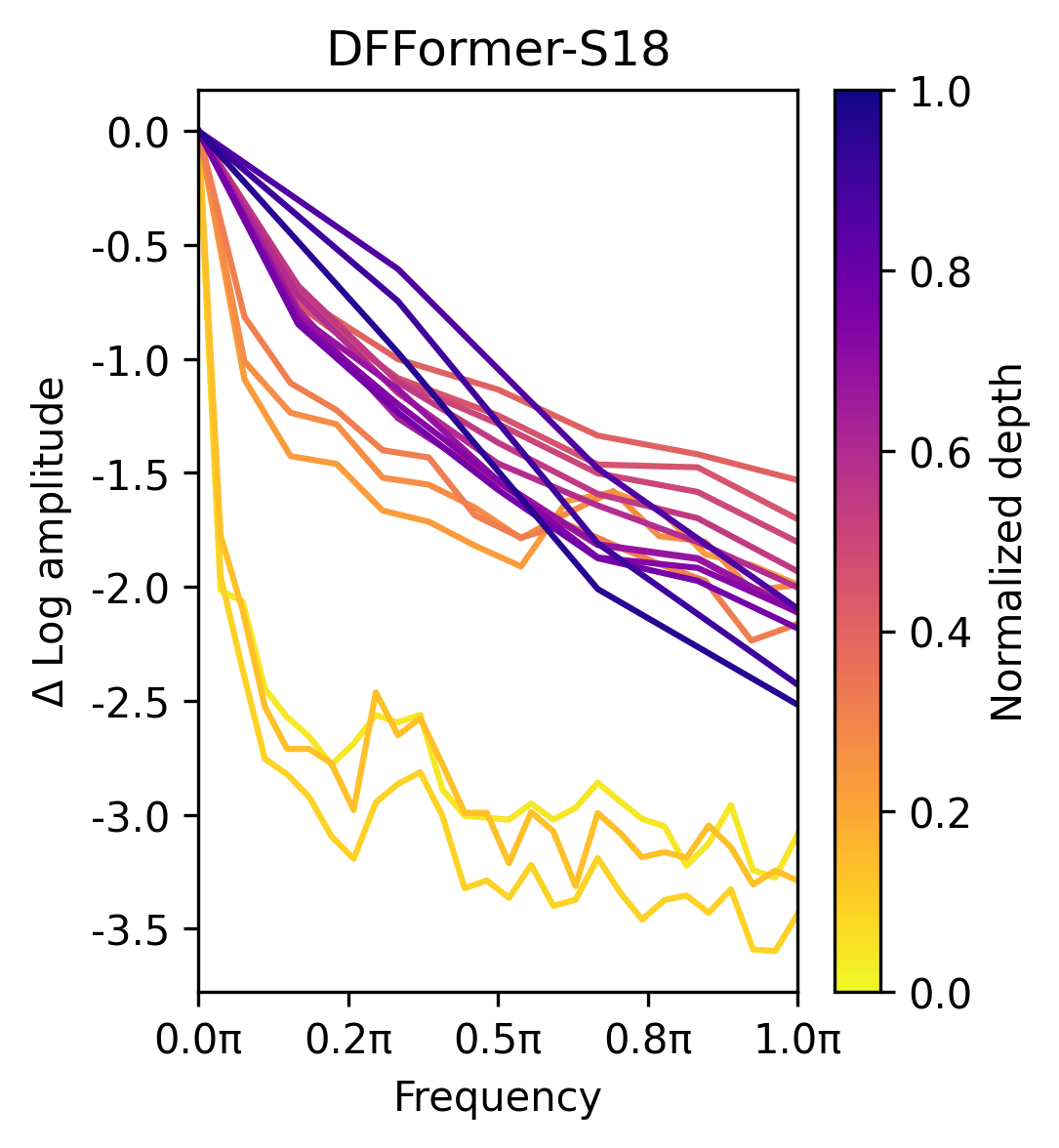}} \quad
\subfloat{\includegraphics[width=0.234\textwidth]{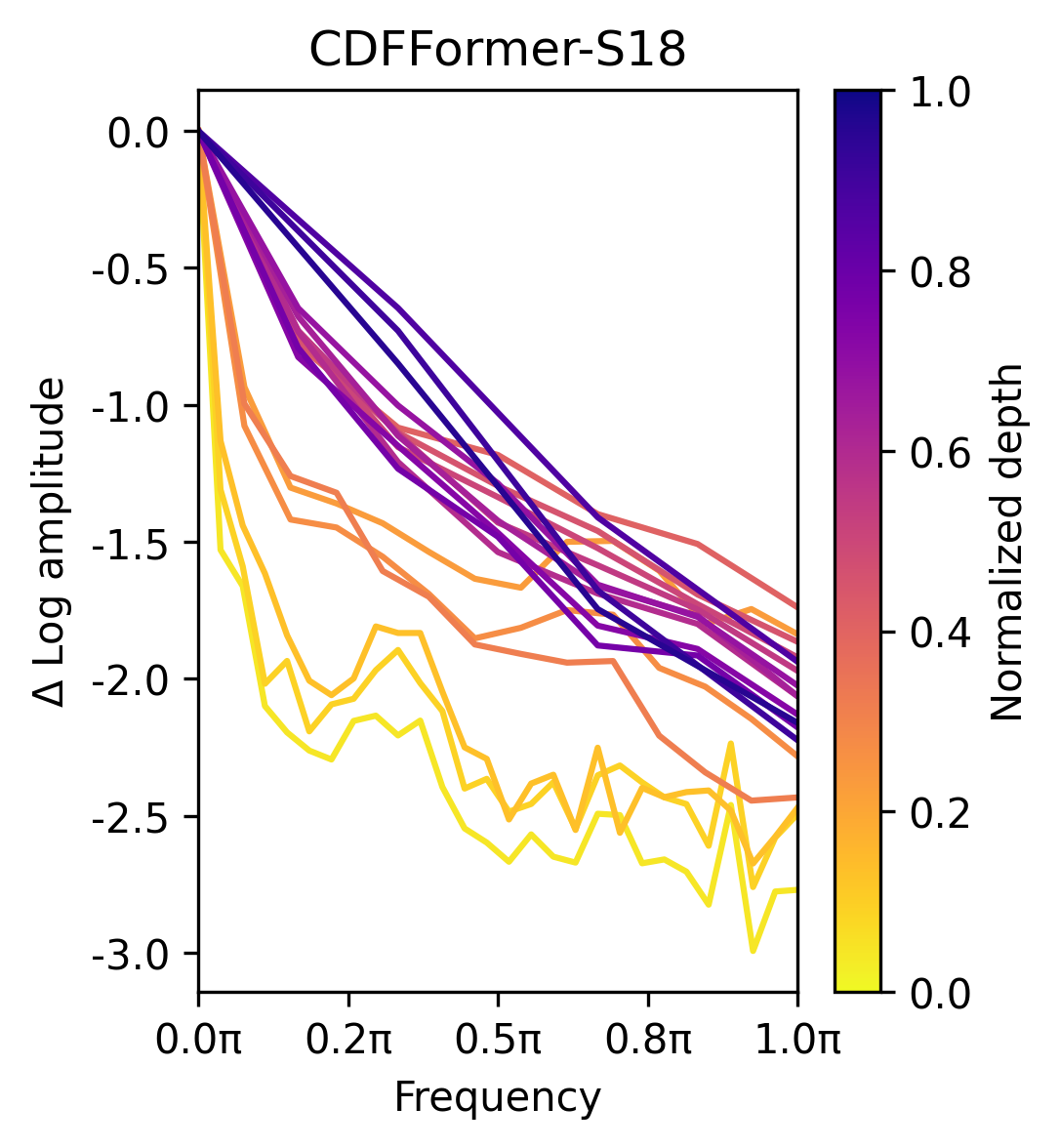}} \\   

\caption{\textbf{Relative log amplitudes of Fourier transformed feature maps on all stages}. All models follow the MetaFormer baseline~\cite{yu2023metaformer} with configurations identical to Table~\ref{table:arch_config}, except for the token mixers and activations.}
\label{fig_relative_log_amp}
\end{figure}

Fig. ~\ref{fig_relative_log_amp} depicts the relative log amplitudes of Fourier transformed features in different blocks for each token mixer. Across all models, high-frequency components remain less constrained (\textit{i.e.}, shallow slope) in the lower layers (bright yellow curves) but undergo increasing suppression (\textit{i.e.}, steep slope) in deeper layers (dense purple curves), leading to a bias toward low-frequency components. Notable differences arise in the middle layers: ConvFormer-S18 shows a gradual decline in high-frequency components, whereas DFFormer-S18 applies mild suppression, slightly attenuating high frequencies. In contrast, CAFormer-S18 and CDFFormer-S18 exhibit an opposite trend. Meanwhile, SPANet-S18-pure and -hybrid demonstrate larger fluctuations in the middle layers, suggesting more adaptive spectral modulation across stages. These findings indicate that token mixers with optimal flexibility in spectral adaptation can enhance vision model performance.

\subsection{Models on Texture and Shape Bias}
\begin{figure}[htbp]
\centering
\subfloat{\includegraphics[width=\linewidth]{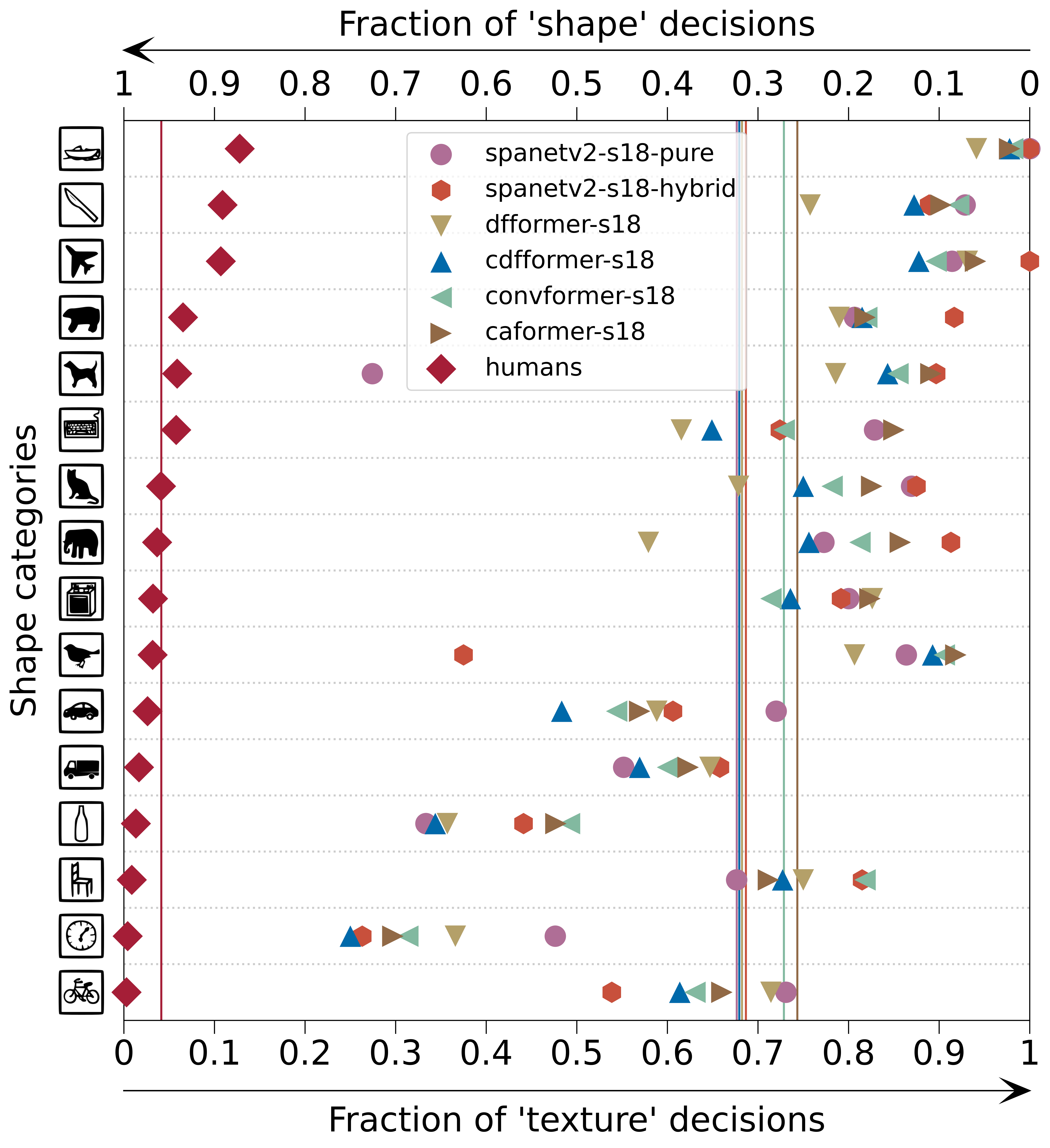}} \\
\subfloat{\includegraphics[width=\linewidth]{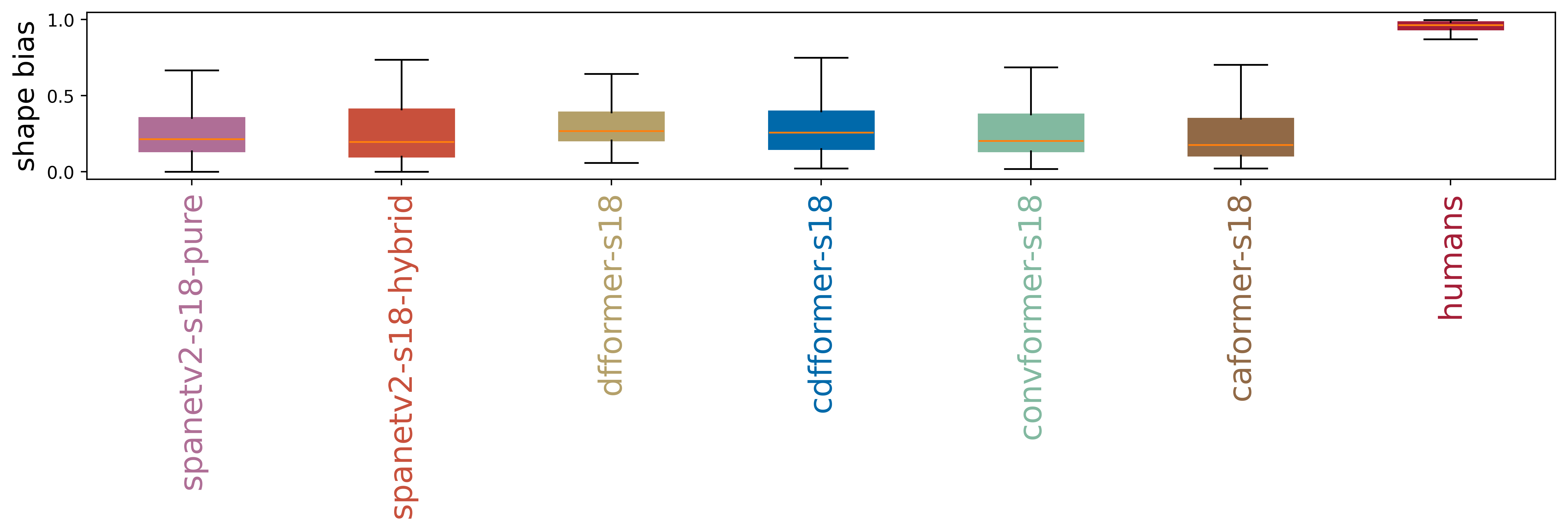}} 
\caption{\textbf{Evaluation of models on texture and shape bias}. All models are pretrained on ImageNet-1K classification using the same augmentations as the MetaFormer baseline~\cite{yu2023metaformer}.}
\label{fig_shape_bias}
\end{figure}

Previous studies suggest that spectral components encode fine details and global structures~\cite{chen2019drop, cooley1969fast, deng1993adaptive}, related to texture and shape bias~\cite{subramanian2023spatial,wang2020high}. To evaluate them, we utilize the \texttt{modelvshuman} toolbox~\cite{geirhos2021partial_code}, with results presented in Fig.~\ref{fig_shape_bias}. The figure illustrates texture–shape bias across various models and human responses, illustrating the extent to which each model depends on texture versus shape cues for recognition. Despite pretraining on ImageNet-1K classification with identical augmentations as the Metaformer baseline~\cite{yu2023metaformer}, models exhibit distinct trends based on their token mixer types. For instance, DFFormer-S18 and CDFFormer-S18, which employ an input-adaptive FFT-based dynamic filter, demonstrate stronger shape bias than ConvFormer-S18 and CAFormer-S18. SPANetV2 models demonstrate comparable texture–shape bias to DFFormer-S18 and CDFFormer-S18, suggesting similar spectral filtering capabilities. However, as shown in Fig.~\ref{fig_relative_log_amp}, SPANetV2 exhibits superior spectral adaptability. Notably, as illustrated in the boxplot below, SPANetV2-S18-hybrid shows a wider min-max range in shape bias than SPANetV2-S18-pure, indicating greater flexibility in spectral adaptation. This adaptability translates into state-of-the-art performance on ImageNet-1K classification and multiple downstream tasks (\textit{i.e.}, object detection, instance segmentation, and semantic segmentation). These findings further highlight the importance of developing spectral-adaptive token mixers to optimize texture-shape bias and improve vision model performance.

\subsection{Complexity and Latency}
As shown in Table~\ref{tab:imagenet-1k_results}, the MetaFormer-based variants exhibit comparable parameter counts and MACs, yet their inference latencies differ noticeably. To investigate the source of this gap, we analyze the computational complexity and latency of each token mixer as a function of the spatial resolution $J = H \times W$ and feature dimension $D$. Since Attention is a shared component across all hybrid architectures, we focus on the token mixers that distinguish these variants and benchmark SPAM against SepConv~\cite{yu2023metaformer} and DynamicFilter~\cite{tatsunami2024fft}.

\subsubsection{Complexity and Parameters}

Table~\ref{table:complexity} summarizes the computational complexity and parameter counts for each token mixer. First, SepConv adopts an inverted bottleneck design with an expansion ratio $\tau$, where the complexity is dominated by two point-wise convolutions ($D \rightarrow \tau D \rightarrow D$), contributing $2\tau JD^2$. The depthwise convolution with a kernel size $m$, utilizing a group-wise structure, adds $\tau JDm^2$, while the activation function contributes $3\tau JD$. This results in an overall complexity of $2\tau JD^2 + \tau JDm^2 + 3\tau JD$ with $2\tau D^2 + \tau Dm^2$ parameters. DynamicFilter similarly incurs $2\tau JD^2$ from its two point-wise projections. However, its spectral-domain processing involves a 2D FFT and inverse FFT applied to the expanded channels ($\tau D$), introducing a complexity term of $2\tau JD \log J$. Additionally, the reweighting MLP, configured with a reweight ratio $\rho$ and number of filters $M$, adds $(\rho + \rho M\tau)D^2$, yielding a total complexity of $2\tau JD^2 + 2\tau JD\log J + (\rho + \rho M\tau)D^2$ with $(2\tau + \rho + \rho M\tau)D^2$ parameters. Finally, SPAM utilizes three linear projections (value, context, and output) contributing $3JD^2$, along with a projection layer after spectral modulation adding $\tau JD^2$. The multi-scale depthwise convolutions, employing $N$ heads with multi-scale kernels where $S_N$ is the sum of squared kernel sizes, contribute $\frac{S_N}{N}JD$, and the spectral rescaling via SRF on $D$ channels adds $2JD \log J$, yielding a total complexity of $(3+\tau)JD^2 + 2JD\log J + \frac{S_N}{N}JD$ with $(3+\tau)D^2 + (\frac{S_N}{N}+\tau(N+2))D$ parameters. Consequently, SPAM maintains a parameter count comparable to that of SepConv (dominant term $(3+\tau)D^2$ vs. $2\tau D^2$). Furthermore, by avoiding the heavy overhead of reweighting MLPs and spectral operations on the expanded channels of DynamicFilter, SPAM achieves a balance between spatial-spectral capability and computational cost.

\subsubsection{Latency Analysis}

\begin{table}    
    \caption{Computational complexity and parameter counts of token mixers.}
    \label{table:complexity}
    \centering
    
    \resizebox{\columnwidth}{!}{
    \begin{tabular}{lcc}
        \toprule
        Method            & Complexity (MACs) &  Parameters   \\ 
        \midrule
        SepConv~\cite{yu2023metaformer}           & $2\tau JD^2 + \tau JDm^2 + 3\tau JD$ & $2\tau D^2 + \tau Dm^2$   \\
        DynamicFilter~\cite{tatsunami2024fft}     & $2\tau JD^2 + 2\tau JD\log J + (\rho+\rho M\tau)D^2$  & $(2\tau + \rho + \rho M\tau)D^2$     \\
        SPAM (ours)       & $(3+\tau)JD^2 + 2JD\log J + \frac{S_N}{N}JD$  & $(3+\tau)D^2 + (\frac{S_N}{N}+\tau(N+2))D$     \\  
        \bottomrule
    \end{tabular}
    }
    
    \begin{tablenotes}[flushleft]
        \footnotesize
        \item[\hskip -\fontdimen 2 \font] Note: $J$: spatial token; $D$: feature dimension; $\tau$: expansion ratio; $m$: kernel size; $M$: number of filters; $\rho$: reweight ratio; $N$: number of heads; $S_N$: sum of squared kernel sizes.
    \end{tablenotes}
\end{table}
\begin{figure}[t] 
    \centering
    \includegraphics[width=\linewidth]{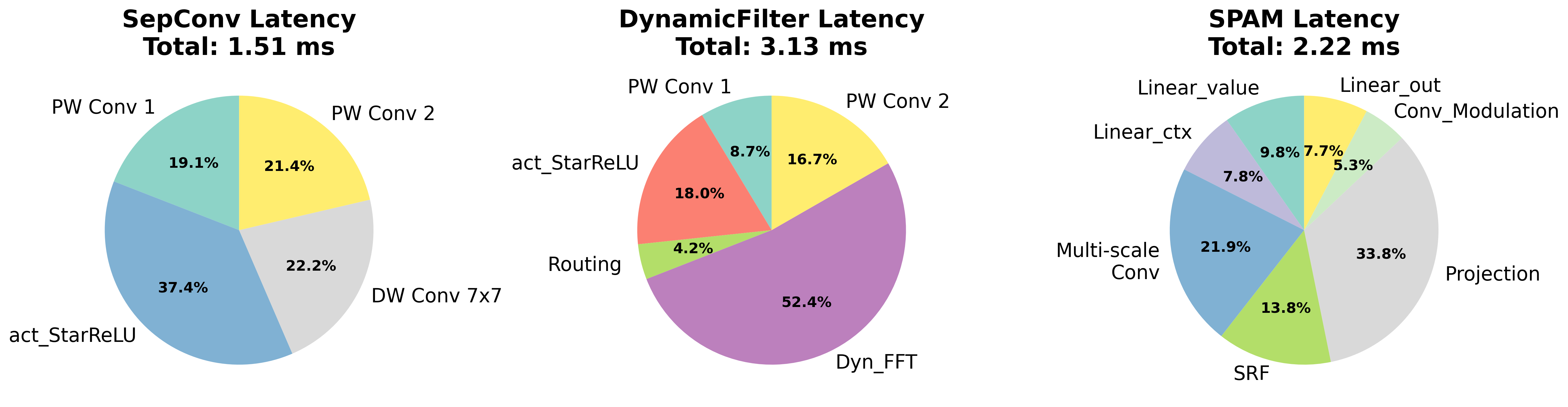} 
    \caption{\textbf{Inference latency breakdown of token-mixer components}. Latency is measured on an NVIDIA RTX 4090 GPU with a batch size of 128. The spatial resolution ($J=196$) and feature dimension ($D=320$) correspond to the Stage-3 input of S18 and S36 models listed in Table ~\ref{table:arch_config}.} 
    \label{fig_latency}
\end{figure}

To empirically validate the complexity analysis, we break down the inference latency of each module in Fig.~\ref{fig_latency}. First, SepConv serves as a purely spatial baseline with the lowest total latency (1.51 ms). Its cost is distributed among standard arithmetic operations, with point-wise convolutions (19.1\% + 21.4\%) and the activation function (act\_StarReLU, 37.4\%) accounting for the majority. While efficient, it lacks spectral modulation. Conversely, DynamicFilter exhibits the highest total latency (3.13 ms), where the spectral operation (Dyn\_FFT) dominates the cost, accounting for 52.4\% of the total time. This high latency stems from its input-adaptive dynamic spectral filtering mechanism applied to the expanded dimension ($\tau D$), which inherently demands heavier spectral modulation computations and complex-valued operations. In contrast, SPAM demonstrates a balanced trade-off with a total latency of 2.22 ms. Notably, the spectral rescaling via SRF accounts for only 13.8\% of the latency. Although SPAM utilizes significantly fewer spectral operations than DynamicFilter by operating on the original feature dimension ($D$), it still achieves optimal flexibility in spectral adaptation, as analyzed in Section~\ref{sec:log-amp}. While the Multi-scale Convolution (21.9\%) and Projection layers (33.8\%) introduce computational overhead, the ablation study in Table~\ref{table:ablation_kernel} confirms their necessity. As discussed in our method design, fixed-size convolutions constrain spectral modulation capabilities. Specifically, relying solely on a small kernel ($3\times3$) reduces latency but limits the capability to capture strong low-frequency global information compared to larger kernels. Conversely, merely increasing the kernel size ($9\times9$) drastically increases latency (131.48 ms) and shifts the spectral behavior to resemble a low-pass filter, thereby diminishing the ability to capture mid- to high-frequency details. Therefore, our multi-scale design offers an optimal solution that overcomes the spectral limitations of fixed-size kernels by facilitating wide-range spectral component aggregation covering a broad spectrum of low and high frequencies. This enables SPAM to achieve superior spectral adaptability and accuracy while avoiding the prohibitive computational burden of DynamicFilter.

\begin{table}[t]
    \caption{\textbf{Impact of kernel scale diversity.} }
    \label{table:ablation_kernel}
    \centering
    \resizebox{\columnwidth}{!}{
    \begin{tabular}{l c c}
        \toprule
        Method (SPANetV2-S18-pure) & Latency (ms) & Top-1 Acc (\%) \\
        \midrule
        Single-scale $3\times3$ & \textbf{94.3} & 83.1 \\  
        Single-scale $5\times5$ & 95.1 & 83.1 \\
        Single-scale $9\times9$ & 131.48& 83.2 \\
        \midrule
        \textbf{Multi-scale (Ours)} & 103.0 & \textbf{83.4} \\
        \bottomrule
    \end{tabular}
    }

\end{table}

\subsection{Qualitative Analysis of Feature Representations}

\begin{figure}[t] 
    \centering
    \setlength{\tabcolsep}{0.5pt} 
    \renewcommand{\arraystretch}{0.5}

    \scriptsize 

    \begin{tabular}{c c c c c}
        \raisebox{3.0em}{\sffamily\tiny hammerhead} & 
        \includegraphics[width=0.22\linewidth]{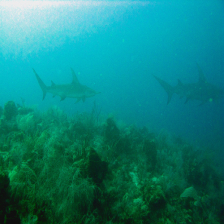} &
        \includegraphics[width=0.22\linewidth]{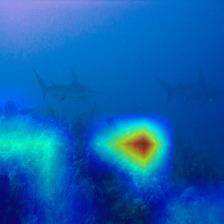} &
        \includegraphics[width=0.22\linewidth]{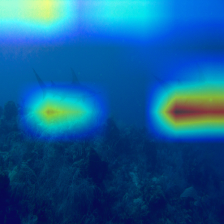} &
        \includegraphics[width=0.22\linewidth]{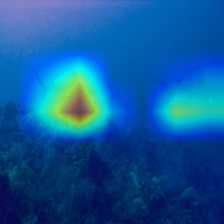} \\

        \raisebox{3.0em}{\sffamily\tiny thresher} & 
        \includegraphics[width=0.22\linewidth]{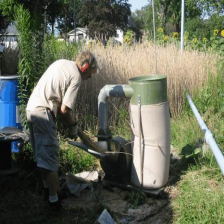} &
        \includegraphics[width=0.22\linewidth]{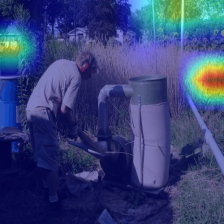} &
        \includegraphics[width=0.22\linewidth]{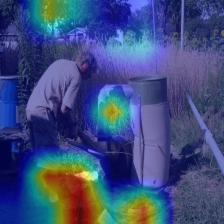} &
        \includegraphics[width=0.22\linewidth]{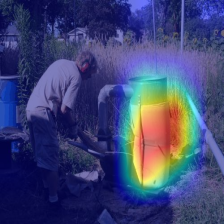} \\

        \raisebox{3.0em}{\sffamily\tiny carbonara} & 
        \includegraphics[width=0.22\linewidth]{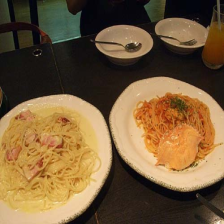} &
        \includegraphics[width=0.22\linewidth]{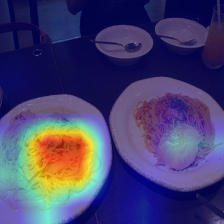} &
        \includegraphics[width=0.22\linewidth]{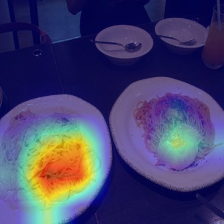} &
        \includegraphics[width=0.22\linewidth]{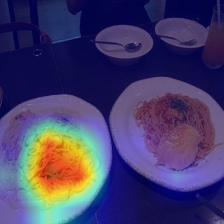} \\

        \raisebox{3.0em}{\sffamily\tiny warplane} & 
        \includegraphics[width=0.22\linewidth]{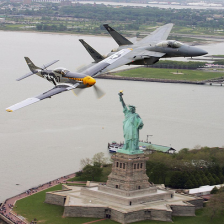} &
        \includegraphics[width=0.22\linewidth]{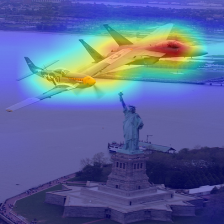} &
        \includegraphics[width=0.22\linewidth]{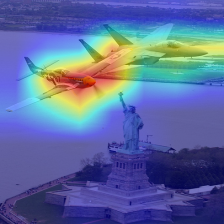} &
        \includegraphics[width=0.22\linewidth]{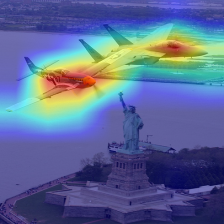} \\
        
         & 
        \tiny input & 
        \tiny ConvFormer-S18 & 
        \tiny DFFormer-S18 & 
        \tiny SPANetV2-S18-pure \\
        
    \end{tabular}
    
    \vspace{2mm}
    \caption{\textbf{Visual comparison of Score-CAM~\cite{wang2020score} activation maps of models trained on ImageNet-1K.} The source images are selected from the validation set.}
    \label{fig:comparison_CAM}
\end{figure}  
Following the complexity analysis of the token mixers, we further visualize the class activation maps (CAM) using Score-CAM~\cite{wang2020score} to intuitively understand how the proposed spectral-adaptive mechanism contributes to representation learning. To directly contrast the behavioral differences of the analyzed token mixers (\textit{i.e.}, SPAM, SepConv, and DynamicFilter), we focus our visualization on their corresponding representative models: SPANetV2-S18-pure, ConvFormer-S18, and DFFormer-S18. Fig.~\ref{fig:comparison_CAM} presents the qualitative comparison of samples from the ImageNet-1K validation set.

As shown in the figure, SPANetV2 generates significantly more precise and distinct activation maps compared to other methods. For the \textit{hammerhead} and \textit{thresher} classes, ConvFormer-S18 fails to locate the target object accurately, instead focusing on background textures such as aquatic vegetation or grass. While DFFormer-S18 captures the target objects, its activation maps are diffuse and contain noticeable noise in non-object regions. In contrast, SPANetV2-S18-pure precisely localizes the semantic region of the true object, effectively suppressing the background clutter. This visually demonstrates that our spectral-adaptive token mixer successfully filters out irrelevant spectral components and focuses on underlying structural boundaries essential for recognition.

Furthermore, in the \textit{carbonara} and \textit{warplane} samples, SPANetV2 demonstrates superior localization capabilities. Specifically, in the \textit{carbonara} image which contains two different pasta dishes, SPANetV2-S18-pure uniquely concentrates its attention exclusively on the target carbonara. In contrast, competing models fail to distinguish the target from the adjacent dish, incorrectly highlighting the non-carbonara pasta as well. Similarly, for the \textit{warplane} image containing multiple instances, our model distinctly highlights the salient regions of both aircraft with higher intensity than competing models. These qualitative results visually corroborate the quantitative performance gains reported in Table~\ref{tab:imagenet-1k_results}. Moreover, the ability to generate object-aligned feature maps directly translates into the superior performance of SPANetV2 in downstream dense prediction tasks, such as object detection and semantic segmentation, as demonstrated in Tables~\ref{table:coco} and~\ref{table:ADE20K}.

\section{Conclusion}
\label{sec:conclusion}
\subsection{Discussion}
In this paper, we conduct a theoretical analysis of the spectral properties of 2D convolution and self-attention through graph spectral analysis. We identify node connectivity, modulated by window size, as a key factor in differentiating their spectral characteristics. Our findings align with prior studies, confirming that self-attention primarily exhibits low-pass filtering behavior, whereas convolution is more effective at capturing mid-/high-frequency components. Additionally, we discuss that larger kernels function as low-pass filters, providing insights into why they favor shape bias—corresponding to low-frequency components that encode shape information~\cite{subramanian2023spatial,ding2022scaling}.

Building on these insights, we propose SPAM, a novel token mixer designed to capture visual features in a spectral-adaptive manner. Specifically, we employ multiple kernels to extract diverse spectral components. Furthermore, inspired by the spectral analysis of 2D convolution, we introduce SRF, a spectral re-scaling filter that adjusts spectral components filtered by convolution. To achieve this, we reformulate the scaling problem as a mask filtering problem using the 2D FFT. With SPAM, we develop a series of SPANetV2 models and evaluate them across three vision tasks, addressing both texture and shape bias. Experimental results demonstrate that SPANetV2 outperforms state-of-the-art models based on convolution, self-attention, and FFT in image classification, object detection and instance segmentation, and semantic segmentation.


\subsection{Limitations}
This study demonstrates that kernel size modulation in 2D convolution induces distinct spectral properties, revealing that large kernels act as low-pass filters akin to self-attention. However, it does not explain why optimizing larger kernels beyond $9\times9$ from scratch is more challenging than optimizing self-attention~\cite{liu2023more,ding2022scaling}. In our experiments, SPAM is evaluated on a limited set of 2D vision tasks. Nonetheless, we believe spectral-adaptive approaches hold significant potential across diverse perception tasks. For instance, extending SPANetV2 as an effective backbone for Vision-Language Models (VLMs) and applying our GSP-based framework to diverse modalities, including audio and non-Euclidean data such as 3D point clouds, remain promising directions for future work. 

Furthermore, while SPAM enhances model performance with a marginal increase in parameters and MACs, its design is not optimized for inference efficiency. Therefore, integrating spectral adaptation with computational efficiency remains an essential challenge. Notably, recent Visual State Space Models (VSSMs)~\cite{zhu2024vision,liu2024vmamba} have demonstrated highly efficient, linear computational complexity, enabling hardware-friendly processing and fast inference speeds. Regarding the accuracy-efficiency trade-off, SPANetV2 achieves superior performance compared to these VSSMs under similar parameter scales, owing to its optimal spectral adaptability. However, it incurs higher inference latency than VSSMs. To bridge this gap, future work could reinterpret their effective scanning strategies as diverse graph structure formulations to embed our spectral-adaptive mechanisms into these highly efficient architectures. We anticipate that this integration could enable the development of an even more capable and adaptable general-purpose vision backbone.



\section*{Acknowledgments}
This work was supported by Samsung Research Funding \& Incubation Center of Samsung Electronics under Project Number SRFC-IT2401-05.







\bibliographystyle{IEEEtran} 
\bibliography{IEEEabrv, egbib}



\vfill

\begin{IEEEbiography}[{\includegraphics[width=1in,height=1.25in,clip,keepaspectratio]{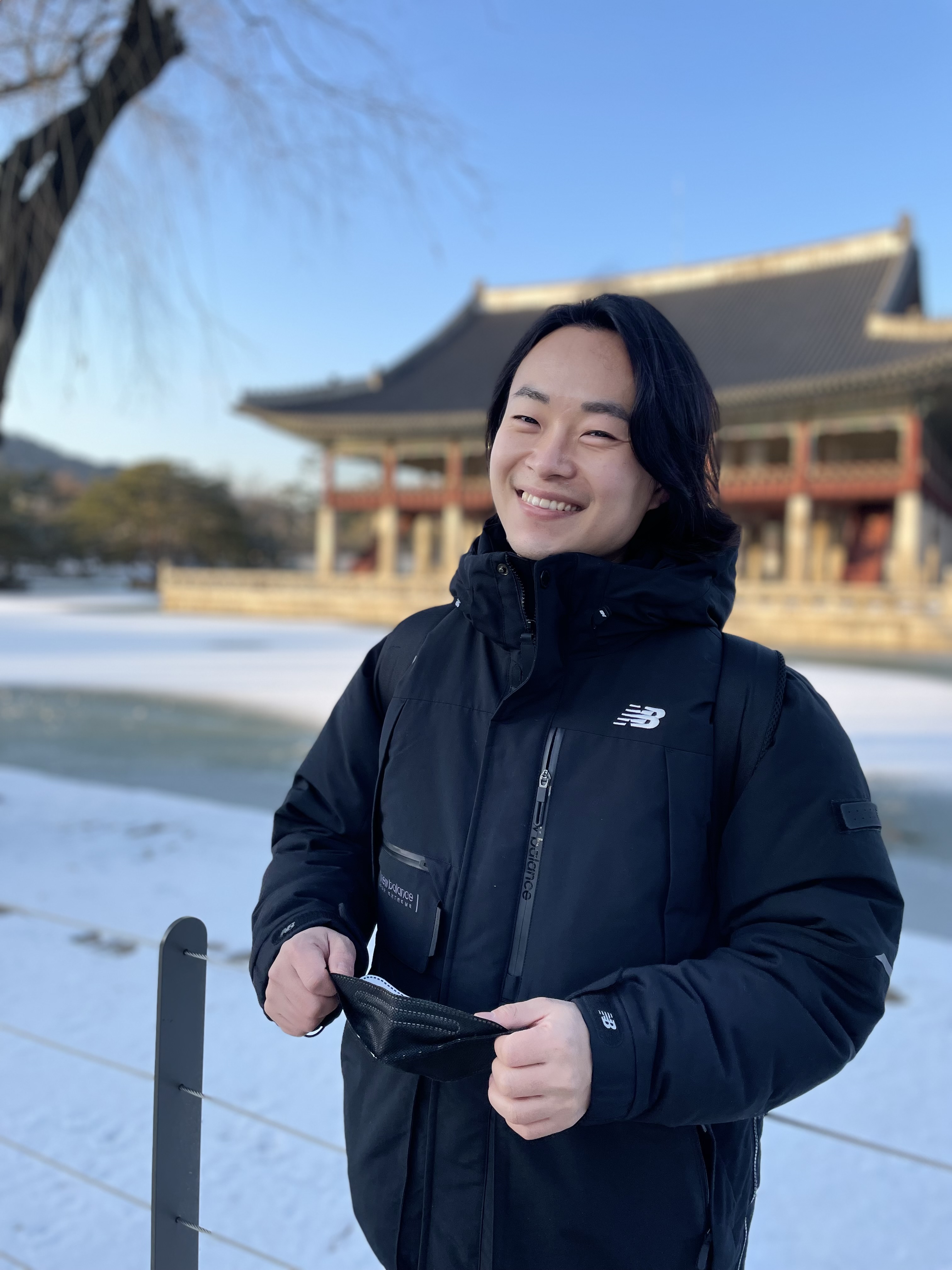}}]{Guhnoo Yun}
received his B.S. degree in Control and Measurement Engineering from Gyeongsang National University (GNU), Jinju, Korea, in 2016, and his M.S. degree in Mechatronics from Gwangju Institute of Science and Technology (GIST), Gwangju, Korea, in 2018. He is currently pursuing a Ph.D. degree in Computer Science and Engineering at Korea University, Seoul, Korea. From 2018 to 2019, he was an intern researcher at the Center for Intelligent Robotics, Korea Institute of Science and Technology (KIST), and he has been a student researcher at the Intelligence and Interaction Research Center, KIST, since 2019. His current research interests are in computer vision, pattern recognition, machine learning, deep learning architecture, and their applications including object detection, instance segmentation, and 3D vision.
\end{IEEEbiography}


\begin{IEEEbiography}
[{\includegraphics[width=1in,height=1.25in,clip,keepaspectratio]{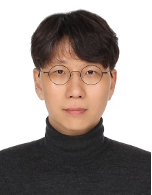}}]{Juhan Yoo}
received the B.S. degree in electronic engineering from Kyunghee University in 2005, the M.S. degree in media engineering from Sogang University in 2009, and the Ph.D. degree in computer science from Korea University, South Korea, in 2019. He was a Senior Researcher at Korea Telecom and Incheon International Airport Corporation, South Korea. He is currently an Assistant Professor in the College of Computer and Artificial Intelligence at Dong-A University, South Korea. His research interests include object detection and image segmentation, robotic vision, pattern recognition, machine learning, and computer graphics.
\end{IEEEbiography}


\begin{IEEEbiography}
[{\includegraphics[width=1in,height=1.25in,clip,keepaspectratio]{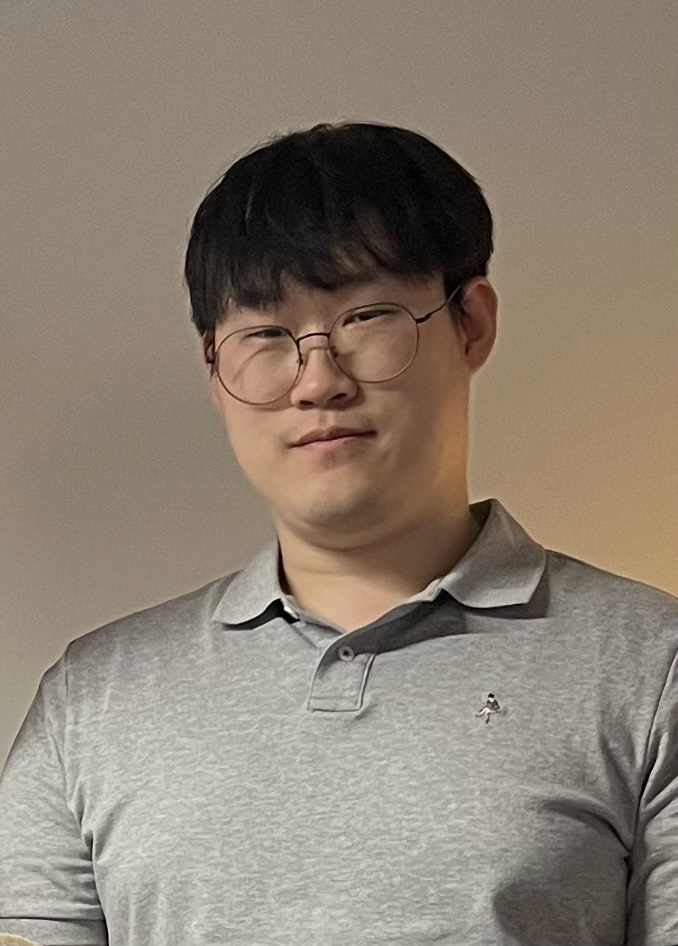}}]{Kijung Kim} received his B.S. and M.S. degrees in Computer Engineering from Kyung Hee University (KHU) in 2012 and 2014, respectively. He is currently pursuing a Ph.D. degree in Computer Science and Engineering at Korea University, Seoul, Korea. From 2016 to 2018, he worked as an intern researcher at the Center for Intelligent Robotics, Korea Institute of Science and Technology (KIST). Since 2019, he has been serving as a student researcher at the Intelligence and Interaction Research Center, KIST. His current research interests include applied areas such as computer vision, machine learning, deep learning architectures, object detection, human pose estimation, and 3D vision.
\end{IEEEbiography}


\begin{IEEEbiography}
[{\includegraphics[width=1in,height=1.25in,clip,keepaspectratio]{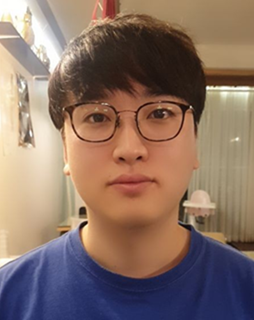}}]{Jeongho Lee} received his B.S. degree in Mechanical System Design Engineering from Seoul National University of Science and Technology (SeoulTech), Seoul, Korea, in 2015.  He then received his M.S. degree in Mechanical Design and Robotics Engineering from SeoulTech in 2017.  From 2017 to 2019, he worked as an intern researcher at the Center for Intelligent Robotics, Korea Institute of Science and Technology (KIST), Seoul, Korea. He is currently pursuing a Ph.D. in Computer Science and Engineering at Korea University, Seoul, Korea, and has been serving as a student researcher at the Intelligence and Interaction Research Center, KIST, since 2019. His current research interests include computer vision, deep learning, and robotics, with applications in object detection, instance segmentation, grasp information generation and object pose estimation.
\end{IEEEbiography}


\begin{IEEEbiography}[{\includegraphics[width=1in,height=1.25in,clip,keepaspectratio]{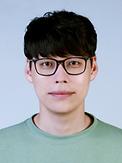}}]{Paul Hongsuck Seo}
received his B.S. degree in Computer Engineering in 2011 from Changwon National University, and his M.S. and Ph.D. degrees in Computer Science and Engineering in 2013 and 2020, respectively, from Pohang University of Science and Technology (POSTECH), South Korea. He is currently an assistant professor at the Department of Computer Science and Engineering of Korea University, Seoul, Korea since 2023. Prior to that from 2020 to 2023, he worked at Google Research as a research scientist. His current research interests are in building multimodal interactive intelligence integrating technologies in computer vision, natural language processing, speech processing and robotics.
\end{IEEEbiography}


\begin{IEEEbiography}[{\includegraphics[width=1in,height=1.25in,clip,keepaspectratio]{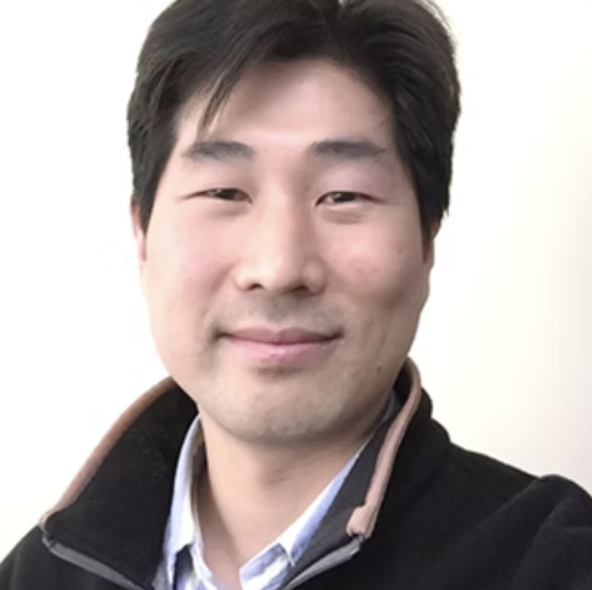}}]{Dong Hwan Kim}
received his B.S. degree in Electrical Engineering in 1999 and his M.S. and Ph.D. degrees in Electrical Engineering and Computer Science in 2001 and 2006, respectively, from Seoul National University (SNU), Seoul, Korea. He is currently a principal researcher at the Intelligence and Interaction Research Center, Korea Institute of Science and Technology (KIST), Seoul, Korea. Additionally, he has been an adjunct professor at Korea University, Seoul, Korea, since 2013. From 2006 to 2007, he worked as a senior engineer at Samsung Electronics, Co., Ltd., Suwon, Korea. In 2007, he joined KIST, and from 2007 to 2010, he was a visiting research associate at the Robotics Institute, Carnegie Mellon University, Pittsburgh, PA, USA. His current research interests are in computer vision, pattern recognition, machine learning, image processing, and their applications including object recognition, image segmentation, object affordance detection, and 3D reconstruction.
\end{IEEEbiography}

\vfill

\end{document}